# Image and Data Mining in Reticular Chemistry Using GPT-4V

Zhiling Zheng,[†,‡,§] Zhiguo He,[†,‡] Omar Khattab,[¶] Nakul Rampal,[†,‡,§] Matei A. Zaharia,[◊] Christian Borgs,[§,◊] Jennifer T. Chayes,[§,◊,○,Δ] and Omar M. Yaghi[†,‡,§,∥],*

[†] Department of Chemistry, University of California, Berkeley, California 94720, United States

[‡] Kavli Energy Nanoscience Institute, University of California, Berkeley, California 94720, United States

[§] Bakar Institute of Digital Materials for the Planet, California 94720, United States

[¶] Department of Computer Science, Stanford University, Stanford, California 94305, United States

[◊] Department of Electrical Engineering and Computer Sciences, University of California, Berkeley, California 94720, United States

[○] Department of Mathematics, University of California, Berkeley, California 94720, United States

[Δ] Department of Statistics, University of California, Berkeley, California 94720, United States

[⊥] School of Information, University of California, Berkeley, California 94720, United States

[∥] KACST−UC Berkeley Center of Excellence for Nanomaterials for Clean Energy Applications, King Abdulaziz City for Science and Technology, Riyadh 11442, Saudi Arabia

*KEYWORDS: materials, data mining, metal–organic frameworks, artificial intelligence, large language models.*

**ABSTRACT:** The integration of artificial intelligence into scientific research has reached a new pinnacle with GPT-4V, a large language model featuring enhanced vision capabilities, accessible through ChatGPT or an API. This study demonstrates the remarkable ability of GPT-4V to navigate and mine complex data for metal-organic frameworks especially from graphical sources. Our approach involved an automated process of converting 346 scholarly articles into 6240 images, which represents a benchmark dataset in this task, followed by deploying GPT-4V to categorize and analyze these images using natural language prompts. This methodology enabled GPT-4V to accurately identify and interpret key plots integral to MOF characterization, such as nitrogen isotherms, PXRD patterns, and TGA curves, among others, with accuracy and recall above 93%. The model's proficiency in extracting critical information from these plots not only underscores its capability in data mining but also highlights its potential in aiding the creation of comprehensive digital databases for reticular chemistry. In addition, the extracted nitrogen isotherm data from those selected literature allowed for a comparison between theoretical and experimental porosity values for over 200 compounds, highlighting certain discrepancies and underscoring the importance of integrating computational and experimental data. This work highlights the potential of AI in accelerating scientific discovery and innovation, bridging the gap between computational tools and experimental research, and paving the way for more efficient, inclusive, and comprehensive scientific inquiry.

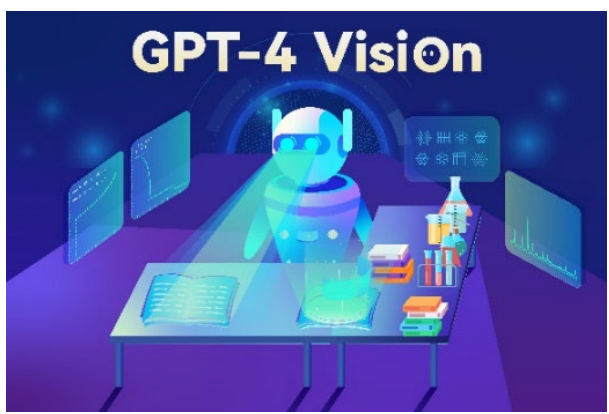

## INTRODUCTION

The integration of artificial intelligence (AI) with chemical sciences holds immense potential, and is accelerated by the rise of large language models (LLMs).[1-7] These models have received significant attention for their remarkable capabilities in assisting with diverse chemistry research tasks and for the fact that they can be intuitively "programmed" or "taught" using daily conversational language.[8-16] It is envisioned that the evolution from text-only to more dynamic, multi-modal LLMs will result in even more powerful and convenient AI assistants across various applications.[5, 17-19]

The recent introduction of GPT-4V, with 'V' denoting its vision capability, stands as a testament to this progress.[17, 20] Trained on a vast and varied collection of multi-modal data, GPT-4V can process and respond to both textual and visual inputs.[20] Its ability to interpret and analyze scientific literature, especially in identifying valuable data within graphical representations, marks a significant advancement over traditional text-only models in natural language processing (NLP).[10, 21-23] These novel capabilities allow researchers from diverse backgrounds, including those with no specialized coding or computer vision expertise, to harness the power of GPT-4V through customized instructions.

Herein, we report the applications of GPT-4V in the field of reticular chemistry.[24, 25] We demonstrate how GPT-4V can seamlessly interpret and integrate complex data from both textual and graphical sources in the scientific literature. This ability to extract and analyze critical information from various figures and plots dramatically increases the model's utility in data mining and information retrieval. In particular, distilling physical characterization results from graphical content reported in papers can be pivotal in establishing databases and guiding the discovery of new chemical compounds.[24, 26-28] Moreover, this methodology, while demonstrated in the context of reticular chemistry, has the potential to be generalized to other scientific disciplines for automated literature reading. This development underscores the expanding role of AI in fostering innovation and discovery, further bridging the gap between advanced computational tools and cutting-edge chemical research.

## RESULTS AND DISCUSSION

**Initial Assessment of GPT-4V's Capabilities.** At the outset of this study, we rigorously evaluated the proficiency of GPT-4V in recognizing and interpreting a range of figures typically found in reticular chemistry literature by asking GPT-4V to describe the respective figure sent along with the chat (Figs. S1 –S14). This assessment encompassed various physical characterization plots including nitrogen isotherms, PXRD patterns, TGA curves, NMR and IR spectra, as well as illustrative plots such as scatter plots, bar plots, and 2D or 3D molecular structures. In addition, synthesis schemes and real experimental images, like microscope and SEM images, were also analyzed. Each figure was presented to GPT-4V alongside a prompt requesting a detailed description.

The responses from GPT-4V (Figs. S1 –S14) demonstrated its remarkable capability to not only categorize these images accurately but also to elaborate on specific details, including notations, axis ranges, color coding and shape of symbols and lines, labels, legends, and to draw inferences from the provided information in the figure caption. This advanced level of contextual data interpretation and holistic analysis underscores the suitability of GPT-4V as a potent AI assistant for image and data mining in scientific literature.

**Designing Prompts for Page Content Labeling.** The core objective of this study was to investigate whether GPT-4V could autonomously navigate through scientific papers, identifying specific information and aggregating it into a comprehensive dataset for further analysis. Our interest specifically centered on plots pivotal in the physical characterization of MOFs. These plots, namely nitrogen isotherms, PXRD patterns, TGA curves, crystal structure or topology illustrations, and other gas sorption isotherms, are instrumental in deducing key properties of chemical compounds, including permanent porosity, crystallinity, thermal stability, connectivity (topology), and sorbent selectivity towards gases. Successfully extracting and consolidating data from these plots amongst a vast volume of literature holds immense potential for advancing our understanding of structure-property relationships and in accelerating the discovery of novel compounds.[8, 24, 27, 28, 30-35]

To this end, we designed specific prompts for GPT-4V, targeting the aforementioned categories (Fig. 1 and ESI Fig. S15). It is noteworthy that these prompts allowed for the possibility of multiple selections on a single page, as it is common for various plots to coexist in scientific literature. Additionally, GPT-4V was instructed to indicate the absence of these five categories when applicable. Therefore, in total six choices were provided for GPT-4V (Fig. 1). The development of these prompts was guided by general principles for prompt engineering in text mining.[10]

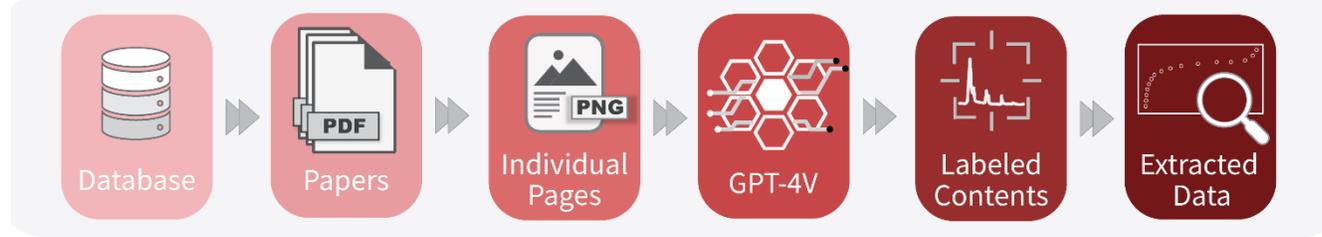

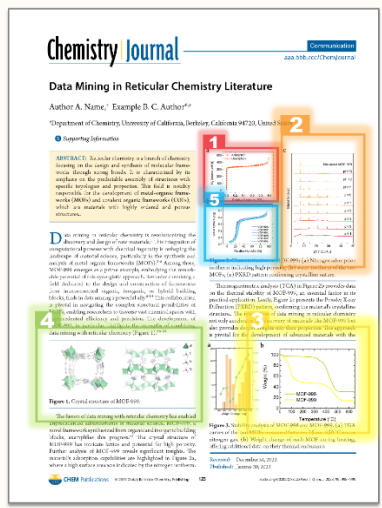
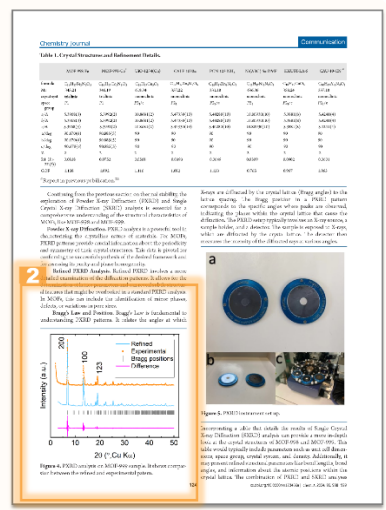
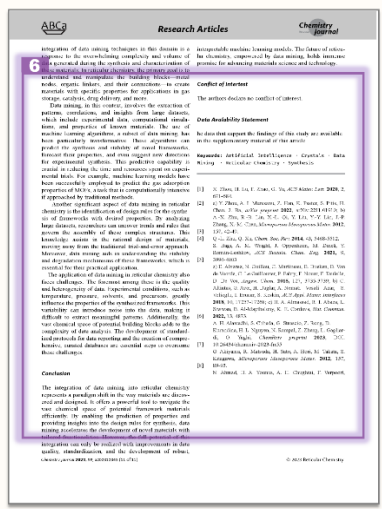
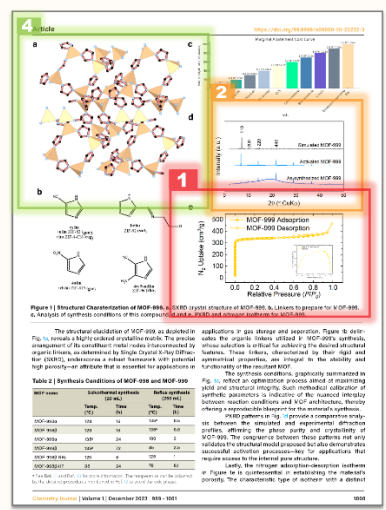

**Fig. 1** Schematic representation of a LLM-driven workflow (top) utilizing image input capabilities for the extraction and categorization of scientific data from chemistry literature. Papers are converted into PNG images and paired with textual prompts to guide GPT-4V in interpreting content within individual pages and responding with appropriate labels. The four example pages depicted are not actual published materials but are non-copyrighted images crafted for illustrative purposes, mimicking the layout of a standard scientific paper with diverse plots and figures (right). The content of these figures is based on actual experimental data, extracted and replotted from earlier publications to accurately reflect the plot shapes for demonstration of the workflow. GPT-4V identifies features based on the described target plots, and the classification responses, which may include single or multiple labels, are compiled (left). These responses are associated with the DOI and corresponding page number of the paper, facilitating efficient data retrieval from targeted pages.

**Performance Evaluation of GPT-4V.** In our workflow, each page of the selected literature was digitized into an image and then analyzed by GPT-4V. This process involved combining each image with a textual prompt, after which GPT-4V's responses were collected. Particularly, GPT-4V was "programmed" by text instruction in the prompt to follow a specific response format (Fig. S15), allowing for the automatic labeling of each page based on its content. While copyright restrictions preclude the sharing of the actual images used in this study, we provide four representative examples that closely simulate the layout and content of the analyzed pages from actual published literature[36-42] to illustrate this figure content identification process (Fig. 1). These examples demonstrate GPT-4V's ability to accurately recognize and label the desired plots on each page, regardless of the complexity of the information shown. Essentially, GPT-4V could perform iterative classifications of each page, utilizing the same prompt but varying the image content (Figs. S17 and S18).

The evaluation of GPT-4V's classification accuracy involved comparing its responses against a benchmarking ground truth dataset. This dataset was meticulously created by experts in reticular chemistry, who manually reviewed and labeled all 6,240 images for the presence of specific content (choices 1 to 6). Performance metrics for each category were calculated individually, considering the multiple-choice nature of the task. For example, if the ground truth for an image was labeled as "1, 3, 4" and GPT-4V's response was "1, 4, 5," the classifications for each category were assessed as follows: nitrogen isotherm (Choice 1) received a True Positive, PXRD pattern (Choice 2) a True Negative, TGA curve (Choice 3) a False Negative, crystal category (Choice 4) a True Positive, other gas sorption isotherm (Choice 5) a False Positive, and "none of the above" (Choice 6) a True Negative.

The summarized metrics are presented in Table 1, displaying promising accuracy rates above 94% for all categories. Precision, recall, and F1 scores varied between 87% and 99% for all categories except for "other gas sorption isotherm." The lower precision in this category is attributed to the category's broad scope not being well instructed in the prompt and occasional mislabeling of IR and NMR spectra as such. This indicates an opportunity for further refinement in prompt specificity. Additionally, similar accuracy rates were observed in GPT-4V's performance via both the web interface and API (Table S1), a testament to the uniformity of the underlying base model. Here we acknowledge that the base model is still in preview version, yet the yielded results are very promising. The automated process for reading papers and collecting output classifications through both interfaces offers diverse operational choices with consistent high performance.

Furthermore, the relation between the ground truth and GPT-4V's classification output was visualized in a confusion matrix (Fig. 2). Among the 6,240 pages analyzed, GPT-4V identified approximately 232 pages with nitrogen isotherms, 630 with PXRD patterns, and 331 with TGA traces. It is important to note that these figures represent the number of pages, not individual plots or compounds, suggesting a substantial volume of data within the analyzed literature. Moreover, 3,537 pages were classified as lacking the plots of interest, which can be a strategy to be applied in the future to help human researchers streamline the review process for a specific type of literature plot by excluding less important pages and focusing only on the rest of the content-rich pages. We also estimated the time of this automated literature processing approach (Tables S2 and S3). On average, processing a 20-page paper took approximately 2 minutes at a rate of 5 seconds per image and this process can be parallelized (Tables S2).

**Table 1 Summary of GPT-4V Performance Metrics in Classifying Scientific Plot Types in Reticular Chemistry Literature.**

| Plot Type | Accuracy | Precision | Recall | F1 Score |
| --- | --- | --- | --- | --- |
| Nitrogen Isotherm | 99.5% | 90.% | 96.7% | 93.5% |
| Power X-Ray Diffraction | 99.2% | 94.3% | 98.4% | 96.3% |
| Thermogravimetric Analysis | 99.2% | 87.8% | 99.3% | 93.2% |
| Crystal Structure or Topology | 98.1% | 93.2% | 97.1% | 95.1% |
| Other Gas Sorption Isotherm | 95.0% | 61.4% | 99.5% | 76.0% |
| None of Above | 94.3% | 96.7% | 93.7% | 95.1% |

**Fig. 2** Confusion matrix displaying the performance of GPT-4V in accurately predicting scientific plot types from reticular chemistry literature. The matrix compares actual choices (ground truth), including Nitrogen Isotherm (NI), Powder X-Ray Diffraction (PXRD), Thermogravimetric Analysis (TGA), Crystal Structure or Topology (CST), Other Gas Sorption Isotherm (OI), and None of the Above (NOA), against the predicted choices made by the GPT-4V with the prompt. The color gradient represents the frequency of predictions, with darker shades indicating higher occurrences.

**GPT-4V's Interpretation of Nitrogen Isotherm Data.** Building upon the successful labeling of page contents, we next directed our focus towards utilizing GPT-4V for the detailed interpretation and analysis of pages featuring nitrogen isotherm plots. To achieve this, we refined our prompt strategy, incorporating additional and specific verbal instructions (Fig. S15). These enhanced prompts guided GPT-4V to not only recognize nitrogen isotherms but also to extract and report key descriptors from each plot. These descriptors included the figure number, compound name, surface area or pore volume value reported by the author, the presence of hysteresis in the adsorption-desorption curve, the saturation plateaus of the isotherm, and an estimation of a bounding box encompassing the figure (Fig. 3). A critical aspect of this approach, aiming to minimize hallucinatory responses, was the instruction for GPT-4V to strictly utilize the information available on the page image, defaulting to "N/A" for any unobtainable data. It turned out that GPT-4V demonstrated a remarkable ability to extract these details by analyzing the isotherm, its axes, legends, and accompanying text content.

To validate the accuracy of GPT-4V's nitrogen isotherm analysis, we manually reviewed over 200 responses from pages containing nitrogen isotherms in our selected papers. This evaluation was conducted for each of the six descriptors independently. The results indicated high accuracy levels for the figure number (96.67%), compound name (90.42%), and porosity analysis (98.33%). We hypothesize that GPT-4V's image processing capabilities, potentially incorporating optical character recognition (OCR) tools such as *pytesseract*, played a pivotal role in these tasks. Given GPT-4V's proficiency with text, it likely excelled in tasks where textual information was directly "readable" from the image. Conversely, the other three descriptors, namely hysteresis presence, saturation plateaus, and bounding box estimation, showed generally satisfactory performance, ranging between 76.25% and 84.58%. These tasks, inherently more challenging and nuanced, required a comprehensive analysis of all image elements.[43] Nonetheless, the overall performance was impressive, especially considering the simplicity with which researchers could instruct GPT-4V using natural language.

**Accelerating Digital Database Creation in Reticular Chemistry.** Our study further demonstrates the potential of GPT-4V in accelerating the establishment of a comprehensive digital database for reticular compounds. This is particularly evident from the experimental measurement outcomes gleaned from community-published literature. Once pages featuring nitrogen isotherm plots are identified, each corresponding curve, typically presented in a non-digital format (e.g., scanned or plotted images), can be meticulously extracted using data extraction tools like WebPlotDigitizer[44]. These extracted data points are then systematically compiled and stored in the database. Fig. 4a presents a collection of nitrogen isotherm data points extracted from the qualified pages in our study as a proof-of-concept, showcasing a diverse array of isotherm types and porosity characteristics.

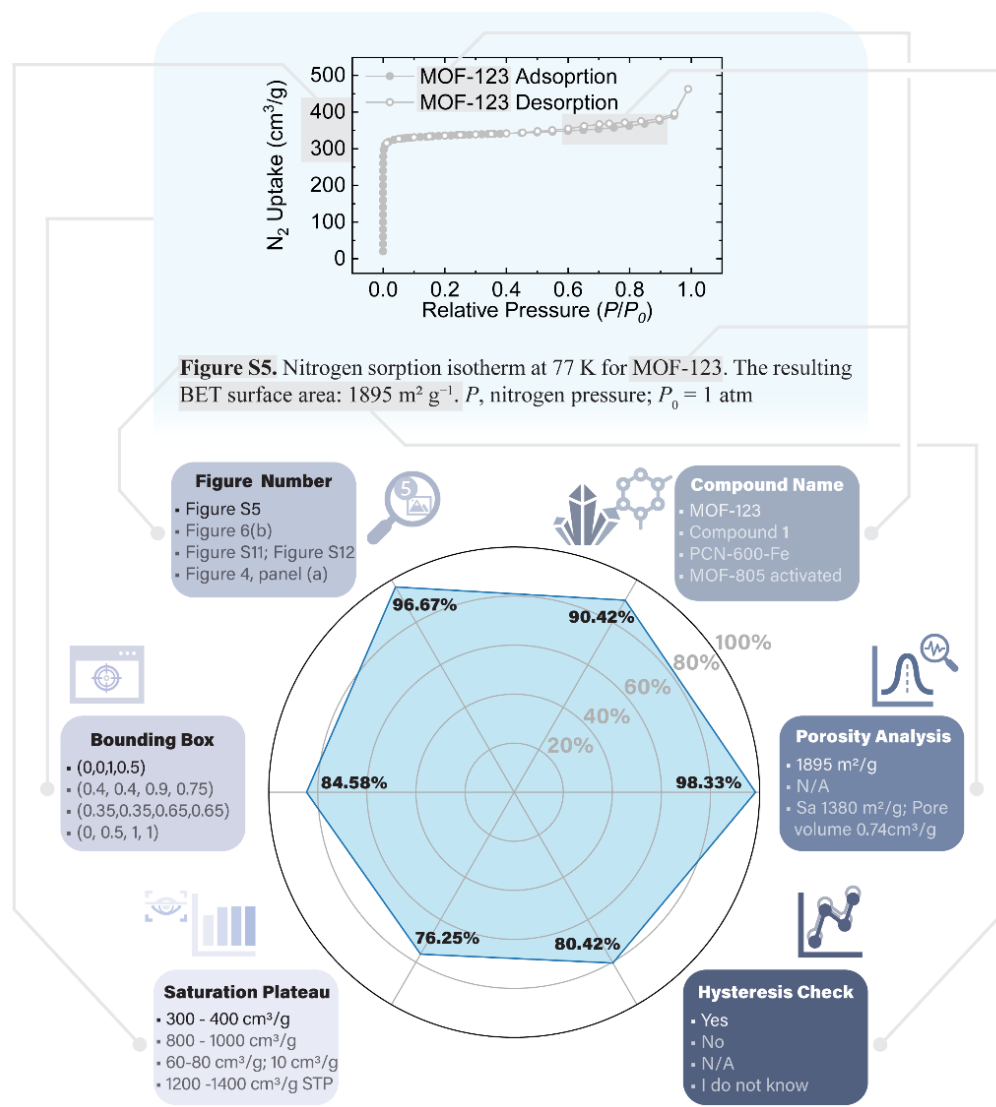

**Fig. 3** The upper section presents an exemplary nitrogen isotherm graph, styled in a manner consistent with standard scientific publications, complete with a representative caption. The lower section features a radar chart detailing the accuracy of GPT-4V's performance in extracting and interpreting critical information, such as figure number, compound name, porosity values, hysteresis presence, saturation uptake, and bounding box parameters. Each vertex on the chart corresponds to the accuracy achieved in its associated category, illustrating the advanced image input capability of GPT-4V in processing and analyzing graphical data from nitrogen isotherms. The grey lines demonstrate textual and visual cues, such as color and shape, that guide GPT-4V in identifying specific information, with sample outcomes for each category displayed within each box.

Leveraging the CoRE MOF Database[26, 29], which provides computational results for compounds discussed in these papers, we matched each adsorption-desorption curve's corresponding compound with their CCDC number, accessible surface area, and pore volume. This enabled us to visualize and compare the experimental porosity (indicated by the nitrogen isotherm curves) with the calculated values (derived from the CoRE MOF Database). The results, illustrated in scatter plots (Figs. 4b and 4c), represent each compound as a data point, where the x-axis denotes the theoretical values calculated from CCDC CIF structures[26, 29], and the y-axis represents the experimentally reported surface area or pore volume inferred from the extracted isotherm. It is crucial to note that our focus was more on the general trends across compounds rather than on pinpointing the exact porosity for each compound, as variations may arise due to differences in calculation models and assumptions.

Interestingly, despite the use of experimentally determined crystal structures in the CoRE MOF Database,[26, 29] discrepancies between theoretical predictions and actual experimental outcomes were observed. For instance, some compounds exhibited a high theoretical porosity based on their refined CCDC structures but failed to demonstrate such porosity experimentally, possibly due to factors like structural collapse during activation, inaccessible pore environments, or suboptimal synthesis conditions. Table 2 presents eight representative examples illustrating these variances, with some compounds showing excellent agreement between theoretical and experimental values (e.g., TAKCAM, OTIHOQ), while others displayed certain deviations (e.g., BOHXED, TAKTAD, and TOCJAY).

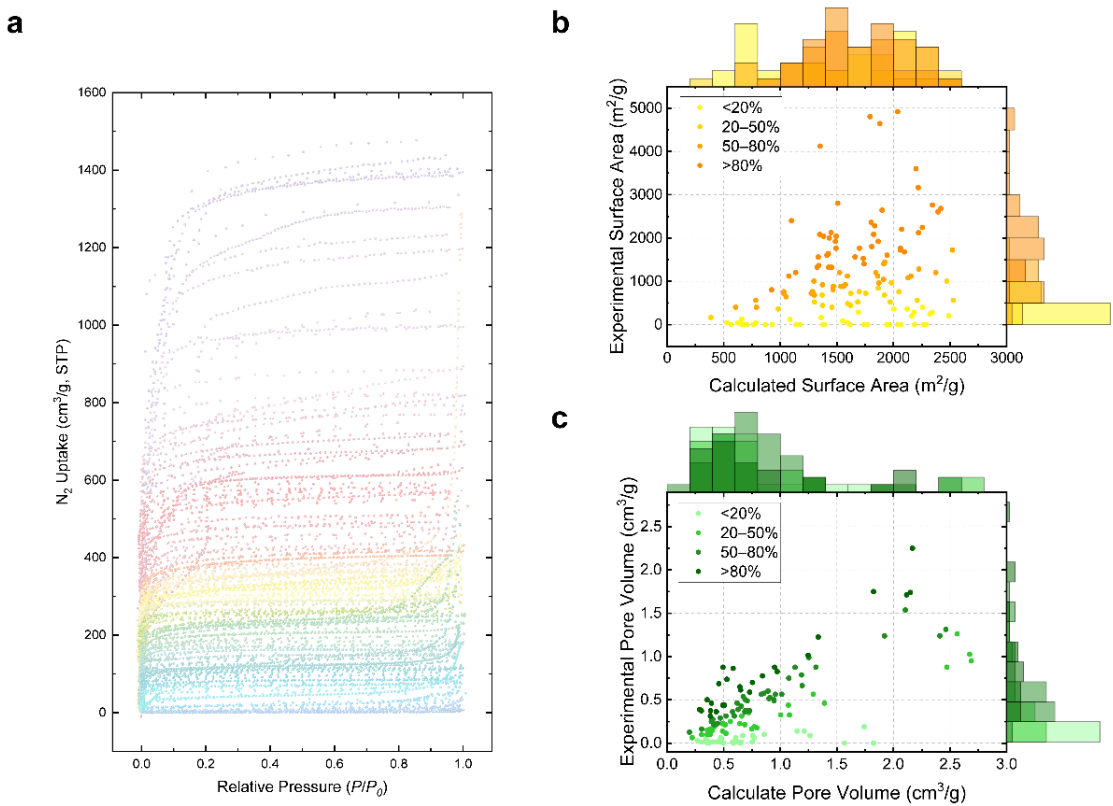

**Fig. 4** Comprehensive analysis of nitrogen isotherm data extracted from selected reticular chemistry literature. (**a**) An overlay scatter plot of distinct adsorption branches for nitrogen isotherms from 347 selected papers, exhibiting a range of uptake behaviors. (**b**) A plot correlating experimental and calculated surface area values, where each point represents one compound's data, linked via the DOI number from the literature to the corresponding CCDC crystal structure in the CoREMOF Database. (**c**) A correlational plot for pore volume, again with each point representing a single compound, illustrating the relationship between experimentally measured values and those calculated based on crystallographic data. The color-coding in (**b**) and (**c**) indicates the percentage of experimental value reaching the theoretical value.

Our findings reveal that reliance solely on computational results for material selection can be sometimes misleading, as many compounds exhibit experimental performances that deviate substantially from theoretical predictions, even when based on experimentally determined structures. It should be note that these experimentally-determined non-porous compounds are not necessarily useless, they in fact serve as valuable negative data points in the mined nitrogen isotherm dataset. By acknowledging these limitations and combining computational methods with experimental data, researchers can gain more comprehensive insights into reticular chemistry, discern trends, and make informed predictions.[45, 46] We envision that leveraging GPT-4V's capabilities to search for more experimental data—not limited to nitrogen isotherm but extending to other isotherms like water, $CO_2$ capture, methane sorption—and other critical plots like TGA curves and PXRD from literature, in tandem with theoretical insights and computational science, can significantly advance the discovery and development of high-performance reticular compounds with desirable properties and functionalities.

To achieve this, it should be acknowledged that the efficiency of using GPT-4V hinges significantly on the skillful crafting of prompts.[10, 43] It has always been a challenge to write good instructions to guide LLMs.[15] Precision in prompt design is essential to achieve specific and accurate results in desirable format, underscoring the need for clear, detailed, and well-articulated instructions. In addition, idiosyncratic and subtle prompt cues can influence the LLMs without changing the underlying meaning, and consequently sometimes equivalent-looking prompts having very different quality. In this context, the emerging tool DSPy (Declarative Self-Improving Python)[47] represents a significant advancement compared with human written prompts. DSPy streamlines the process of prompt creation, combining techniques for both prompting and fine-tuning language models. Applying DSPy for prompt optimization necessitates the existence of a relatively small development set and some data annotation that DSPy will use to conduct the optimization automatically. Conceptually, this innovative system could enable researchers to commence their inquiries with a basic scaffolding of the high-level steps of the design and use DSPy to automate the instruction of LLMs, which can be considered a solution to further strengthen our workflow and improve the performance of LLMs. In the near future, we envision this combined approach can further increase the accuracy of the tasks demonstrated in this study, expand scope of the type of data to be mined from the literature, and hold the promise of making advanced natural language processing tools more accessible and adaptable to diverse research scenarios.

**Table 2** Calculated and Experimental Accessible Surface Areas for Selected Compounds in Reticular Chemistry Literature

| Representative compound | Metal | CCDC code | Surface area ($cm^3/g$) | | Reference |
|---|---|---|---|---|---|
| | | | Calculated[a] | Experimental[b] | |
| ZJU-70a | Cu | DORNAB | 2520 | 1790 | 48 |
| MOF-802 | Zr | BOHXED | 2070 | <10 | 49 |
| Hf-L7 | Hf | TAKCAM | 2070 | 2200 | 50 |
| MFM-300(Ga$_2$) | Ga | TAKTAD | 1600 | 480 | 51 |
| MIL-100 (Al) | Al | BUSPIP | 1350 | 2090 | 52 |
| [Cd(L4)]·1.5DMF | Cd | OTIHOQ | 1330 | 1320 | 53 |
| bio-MOF-101 | Zn | TOCJAY | 1100 | 2400[c] | 54 |
| {[Ni$_6$(N$_3$)$_{12}$L$_6$]·(H$_2$O)$_{13}$}$_\infty$ | Ni | NUZCER | 790 | 400 | 55 |

Note: (a) Based on the accessible surface area calculated for each compound's crystal structure as reported in the CoRE MOF 2019 database[29]. (b) Based on data extracted from the experimental nitrogen isotherm reported in the respective literature, results are rounded to the nearest ten. (c) The reported surface area by the author in the literature is 4410 $m^2/g$.

## CONCLUDING REMARKS

This study demonstrated the application of GPT-4V in text, image, and data mining within the realm of reticular chemistry. We have showcased GPT-4V's capability to process page images alongside specifically crafted prompts, effectively classifying content and pinpointing pages containing desired plots for detailed analysis. A significant insight from our research is the potential generalizability of this approach beyond reticular chemistry to a broader spectrum of scientific disciplines. The key features of LLMs like GPT-4V—being "programmable" through everyday natural language and possessing pre-trained domain knowledge—eliminate traditional barriers associated with coding expertise and model training for specific plot or figure recognition. This adaptability is highlighted by the ease of transitioning from analyzing one type of data, such as TGA curves, to another, like water isotherms, simply by modifying the instructional prompt. Furthermore, we envision that to make the instruction on GPT-4V more effective, the integration of advanced platform like DSPy into our workflow opens new avenues in scientific data mining and points to a future where AI becomes an integral, user-friendly tool in the advancement of scientific knowledge.

## METHODS

**Data Preparation.** For this study, a comprehensive dataset comprising 6,240 pages from 346 scholarly articles published in 22 peer-reviewed journals among 5 different publishers discussing metal-organic frameworks (MOFs) was meticulously curated from the CoRE MOF Database (2019)[29] to ensure the diversity of writing styles, article formatting and page layout. The selection criteria focused on excluding papers reporting structures zero accessible surface areas and those without publicly available DOI numbers. The selected PDFs were transformed into PNG format utilizing a Python script developed with the assistance of ChatGPT (Fig. S16). This conversion process yielded 4423 images from the main text and 1817 from supporting information of those chosen papers (ESI Dataset).

**Prompts and Image Mining Considerations.** We wrote instructions in natural language to guide GPT-4V for image classification and analysis. The primary objective was to enable GPT-4V to categorize each image with a label reflecting its content. These labels were designed to represent the most prevalent types of plots encountered in reticular chemistry, encompassing essential data and values. Key plots include the nitrogen adsorption-desorption isotherm (indicating porosity), the PXRD pattern (illustrating crystallinity), the TGA curve (demonstrating thermal stability), crystallographic structure rendering images (depicting crystal structure or topology), and other gas sorption isotherms (relevant for applications and gas uptake). Accordingly, these were established as label choices 1 through 6. It is noteworthy that a single page may present multiple plots (e.g., PXRD and TGA concurrently), necessitating GPT-4V's capability to perform an identification of multiple formats. The primary task assigned to GPT-4V involves identifying these specified elements within the image of the whole page, complemented by a structured response template. Furthermore, the crafting of prompts adhered to three guiding principles outlined in our previous publication[10]: (i) minimizing hallucination, (ii) implementing detailed instructions, and (iii) ensuring structured output. The complete prompt is available in ESI Fig. S15.

**Automation in Paper Reading.** With the prompt finalized, GPT-4V's task was to "read" through all the pages from selected papers. This was accomplished by sequentially presenting each image of the whole page, along with the designated prompt, to the model and collecting its responses. This iterative process was automated using a for-loop structure. In particular, two options exist, and both were tested in this study: (i) interfacing with the web-based chatbot (ChatGPT powered by GPT-4V, version dated September 25, 2023) and (ii) utilizing an API to connect with the *gpt-4-1106-vision-preview* model. Both methods were facilitated through Python scripts capable of operating autonomously ([Figs. S17](#) and [S18](#)). Additionally, the process for extracting responses from GPT-4V was automated by Python script ([Figs. S19](#) and [S20](#)).

## ASSOCIATED CONTENT

**Supporting Information**. Full prompts designed to guide GPT-4V; Additional examples showcasing GPT-4V's performance in reading various figure inputs and its corresponding responses; Python code used to automate the data mining and analysis processes; Detailed information on the selected papers in this study is provided, including the ground truth and the classification output for each page in a spreadsheet format; Tabulated data for all extracted nitrogen isotherms in this study.

## AUTHOR INFORMATION

### Corresponding Author

**Omar M. Yaghi** – *Department of Chemistry; Kavli Energy Nanoscience Institute; and Bakar Institute of Digital Materials for the Planet, University of California, Berkeley, California 94720, United States; UC Berkeley–KACST Joint Center of Excellence for Nanomaterials for Clean Energy Applications, King Abdulaziz City for Science and Technology, Riyadh 11442, Saudi Arabia;* orcid.org/0000-0002-5611-3325; Email: yaghi@berkeley.edu

### Other Authors

**Zhiling Zheng** – *Department of Chemistry; Kavli Energy Nanoscience Institute; and Bakar Institute of Digital Materials for the Planet, University of California, Berkeley, California 94720, United States;* orcid.org/0000-0001-6090-2258

**Zhiguo He** – *Department of Chemistry and Kavli Energy Nanoscience Institute, University of California, Berkeley, California 94720, United States*

**Omar Khattab** – *Department of Computer Science, Stanford University, Stanford, California 94305, United States*

**Nakul Rampal** – *Department of Chemistry; Kavli Energy Nanoscience Institute; and Bakar Institute of Digital Materials for the Planet, University of California, Berkeley, California 94720, United States;* orcid.org/0000-0002-6187-5631

**Matei A. Zaharia** – *Department of Electrical Engineering and Computer Sciences, University of California, Berkeley, California 94720, United States;* orcid.org/0000-0002-7547-7204

**Christian Borgs** – *Bakar Institute of Digital Materials for the Planet; and Department of Electrical Engineering and Computer Sciences, University of California, Berkeley, California 94720, United States;* orcid.org/0000-0001-5653-0498

**Jennifer T. Chayes** – *Bakar Institute of Digital Materials for the Planet; Department of Electrical Engineering and Computer Sciences; Department of Mathematics; Department of Statistics; and School of Information, University of California, Berkeley, California 94720, United States;* orcid.org/0000-0003-4020-8618

### Author Contributions

Z.Z. and O.M.Y conceived and designed the experiments. Z.Z. and Z.H. performed the data curation. Z.Z. conducted the investigation and developed the methodology. Z.Z. and O.M.Y. carried out the formal analysis and visualization. O.M.Y. supervised the project. Z.Z. wrote the initial draft of the paper. All authors contributed significantly and collaboratively to the data analysis, and the review and editing of the manuscript.

### Notes

The authors declare that they have no conflict of interest.

## ACKNOWLEDGMENTS

This material is based upon work supported by the Defense Advanced Research Projects Agency (DARPA) under contract HR0011-21-C-0020. Any opinions, findings, and conclusions or recommendations expressed in this material are those of the authors and do not necessarily reflect the views of DARPA. Z.Z. expresses gratitude to Mr. Kefan Dong (Stanford University), Mr. Jiayi Weng (OpenAI), Ms. Oufan Zhang (UC Berkeley), Mr. Zichao Rong (UC Berkeley) and Ms. Zeqi Gu (Cornell University) for their valuable discussions. Z.Z. also acknowledges financial support from a Kavli ENSI Graduate Student Fellowship.

## REFERENCES

(1) Birhane, A.; Kasirzadeh, A.; Leslie, D.; Wachter, S., Science in the age of large language models. *Nat. Rev. Phys.* **2023,** *5* (5), 277-280.
(2) White, A. D., The future of chemistry is language. *Nat. Rev. Chem.* **2023,** *7*, 457–458.
(3) Bubeck, S.; Chandrasekaran, V.; Eldan, R.; Gehrke, J.; Horvitz, E.; Kamar, E.; Lee, P.; Lee, Y. T.; Li, Y.; Lundberg, S. Sparks of artificial general intelligence: Early experiments with gpt-4. *arXiv* DOI: 10.48550/arXiv:2303.12712 (accessed 2023-04-13).
(4) AI4Science, M. R.; Quantum, M. A. The Impact of Large Language Models on Scientific Discovery: a Preliminary Study using GPT-4. *arXiv preprint* DOI: 10.48550/arXiv.2311.07361 (accessed 2023-11-13).
(5) OpenAI GPT-4 technical report. DOI: 10.48550/arXiv:2303.08774v3 (accessed 2023-03-27).
(6) Anil, R.; Dai, A. M.; Firat, O.; Johnson, M.; Lepikhin, D.; Passos, A.; Shakeri, S.; Taropa, E.; Bailey, P.; Chen, Z. Palm 2 technical report. *arXiv preprint* DOI: 10.48550/arXiv.2305.10403 (accessed 2023-09-13).
(7) Touvron, H.; Martin, L.; Stone, K.; Albert, P.; Almahairi, A.; Babaei, Y.; Bashlykov, N.; Batra, S.; Bhargava, P.; Bhosale, S. Llama 2: Open foundation and fine-tuned chat models. *arXiv preprint* DOI: 10.48550/arXiv.2307.09288 (accessed 2023-07-19).

(8) Zheng, Z.; Zhang, O.; Nguyen, H. L.; Rampal, N.; Alawadhi, A. H.; Rong, Z.; Head-Gordon, T.; Borgs, C.; Chayes, J. T.; Yaghi, O. M., ChatGPT Research Group for Optimizing the Crystallinity of MOFs and COFs. *ACS Cent. Sci.* **2023**, *9* (11), 2161–2170.
(9) Jablonka, K. M.; Ai, Q.; Al-Feghali, A.; Badhwar, S.; Bocarsly, J. D.; Bran, A. M.; Bringuier, S.; Brinson, L. C.; Choudhary, K.; Circi, D., 14 examples of how LLMs can transform materials science and chemistry: a reflection on a large language model hackathon. *Digital Discovery* **2023**, *2* (5), 1233-1250.
(10) Zheng, Z.; Zhang, O.; Borgs, C.; Chayes, J. T.; Yaghi, O. M., ChatGPT Chemistry Assistant for Text Mining and Prediction of MOF Synthesis. *J. Am. Chem. Soc.* **2023**, *145* (32), 18048–18062.
(11) Kang, Y.; Kim, J. ChatMOF: An Autonomous AI System for Predicting and Generating Metal-Organic Frameworks. arXiv:2308.01423.
(12) Liu, S.; Wang, J.; Yang, Y.; Wang, C.; Liu, L.; Guo, H.; Xiao, C. ChatGPT-powered Conversational Drug Editing Using Retrieval and Domain Feedback. DOI: 10.48550/arXiv:2305.18090 (accessed 2023-05-29).
(13) Bran, A. M.; Cox, S.; White, A. D.; Schwaller, P. ChemCrow: Augmenting large-language models with chemistry tools. arXiv:2304.05376.
(14) Zheng, Z.; Rong, Z.; Rampal, N.; Borgs, C.; Chayes, J. T.; Yaghi, O. M., A GPT-4 Reticular Chemist for Guiding MOF Discovery. *Angew. Chem. Int. Ed.* **2023**, *62*, e202311983.
(15) Nori, H.; Lee, Y. T.; Zhang, S.; Carignan, D.; Edgar, R.; Fusi, N.; King, N.; Larson, J.; Li, Y.; Liu, W. Can Generalist Foundation Models Outcompete Special-Purpose Tuning? Case Study in Medicine. *arXiv preprint* DOI: 10.48550/arXiv.2311.16452 (accessed 2023-11-28).
(16) Zheng, Z.; Alawadhi, A.; Chheda, S.; Neumann, S. E.; Rampal, N.; Liu, S.; Nguyen, H. L.; Lin, Y.-h.; Rong, Z.; Siepmann, J.; Gagliardi, L.; Anandkumar, A.; Borgs, C.; Chayes, J. T.; Yaghi, O., Shaping the Water Harvesting Behavior of Metal-Organic Frameworks Aided by Fine-Tuned GPT Models. *J. Am. Chem. Soc.* **2023**.
(17) Yang, Z.; Li, L.; Lin, K.; Wang, J.; Lin, C.-C.; Liu, Z.; Wang, L. The dawn of lmms: Preliminary explorations with gpt-4v (ision). *arXiv preprint* DOI: 10.48550/arXiv.2309.17421 (accessed 2023-10-11).
(18) Yan, Z.; Zhang, K.; Zhou, R.; He, L.; Li, X.; Sun, L. Multimodal ChatGPT for Medical Applications: an Experimental Study of GPT-4V. *arXiv preprint* DOI: 10.48550/arXiv.2310.19061 (accessed 2023-10-29).
(19) Wu, C.; Yin, S.; Qi, W.; Wang, X.; Tang, Z.; Duan, N. Visual chatgpt: Talking, drawing and editing with visual foundation models. *arXiv preprint* DOI: 10.48550/arXiv.2303.04671 (accessed 2023-03-08).
(20) OpenAI GPT-4V(ision) System Card. *OpenAI Website* https://cdn.openai.com/papers/GPTV_System_Card.pdf (accessed 2023-09-25).
(21) Park, S.; Kim, B.; Choi, S.; Boyd, P. G.; Smit, B.; Kim, J., Text mining metal–organic framework papers. *J. Chem. Inf. Model.* **2018**, *58* (2), 244-251.
(22) Park, H.; Kang, Y.; Choe, W.; Kim, J., Mining Insights on Metal–Organic Framework Synthesis from Scientific Literature Texts. *J. Chem. Inf. Model.* **2022**, *62* (5), 1190-1198.
(23) Luo, Y.; Bag, S.; Zaremba, O.; Cierpka, A.; Andreo, J.; Wuttke, S.; Friederich, P.; Tsotsalas, M., MOF synthesis prediction enabled by automatic data mining and machine learning. *Angew. Chem. Int. Ed.* **2022**, *61* (19), e202200242.
(24) Lyu, H.; Ji, Z.; Wuttke, S.; Yaghi, O. M., Digital reticular chemistry. *Chem* **2020**, *6* (9), 2219-2241.
(25) Yaghi, O. M.; O'Keeffe, M.; Ockwig, N. W.; Chae, H. K.; Eddaoudi, M.; Kim, J., Reticular synthesis and the design of new materials. *Nature* **2003**, *423* (6941), 705-714.
(26) Chung, Y. G.; Camp, J.; Haranczyk, M.; Sikora, B. J.; Bury, W.; Krungleviciute, V.; Yildirim, T.; Farha, O. K.; Sholl, D. S.; Snurr, R. Q., Computation-ready, experimental metal–organic frameworks: A tool to enable high-throughput screening of nanoporous crystals. *Chem. Mater.* **2014**, *26* (21), 6185-6192.
(27) Rosen, A. S.; Fung, V.; Huck, P.; O'Donnell, C. T.; Horton, M. K.; Truhlar, D. G.; Persson, K. A.; Notestein, J. M.; Snurr, R. Q., High-throughput predictions of metal–organic framework electronic properties: theoretical challenges, graph neural networks, and data exploration. *Npj Comput. Mater.* **2022**, *8* (1), 112.
(28) Kancharlapalli, S.; Snurr, R. Q., High-Throughput Screening of the CoRE-MOF-2019 Database for CO2 Capture from Wet Flue Gas: A Multi-Scale Modeling Strategy. *ACS Appl. Mater. Interfaces* **2023**.
(29) Chung, Y. G.; Haldoupis, E.; Bucior, B. J.; Haranczyk, M.; Lee, S.; Zhang, H.; Vogiatzis, K. D.; Milisavljevic, M.; Ling, S.; Camp, J. S., Advances, updates, and analytics for the computation-ready, experimental metal–organic framework database: CoRE MOF 2019. *J. Chem. Eng. Data* **2019**, *64* (12), 5985-5998.
(30) Nandy, A.; Terrones, G.; Arunachalam, N.; Duan, C.; Kastner, D. W.; Kulik, H. J., MOFSimplify, machine learning models with extracted stability data of three thousand metal–organic frameworks. *Sci. Data* **2022**, *9* (1), 74.
(31) Moosavi, S. M.; Nandy, A.; Jablonka, K. M.; Ongari, D.; Janet, J. P.; Boyd, P. G.; Lee, Y.; Smit, B.; Kulik, H. J., Understanding the diversity of the metal-organic framework ecosystem. *Nat. Commun.* **2020**, *11* (1), 1-10.
(32) Nandy, A.; Yue, S.; Oh, C.; Duan, C.; Terrones, G. G.; Chung, Y. G.; Kulik, H. J., A database of ultrastable MOFs reassembled from stable fragments with machine learning models. *Matter* **2023**, *6* (5), 1585-1603.
(33) Nandy, A.; Duan, C.; Kulik, H. J., Using machine learning and data mining to leverage community knowledge for the engineering of stable metal–organic frameworks. *J. Am. Chem. Soc.* **2021**, *143* (42), 17535-17547.
(34) Tan, J. C.; Bennett, T. D.; Cheetham, A. K., Chemical structure, network topology, and porosity effects on the mechanical properties of Zeolitic Imidazolate Frameworks. *Proc. Natl. Acad. Sci. U.S.A.* **2010**, *107* (22), 9938-9943.
(35) Batra, R.; Chen, C.; Evans, T. G.; Walton, K. S.; Ramprasad, R., Prediction of water stability of metal–organic frameworks using machine learning. *Nat. Mach.* **2020**, *2* (11), 704-710.
(36) Wang, B.; Lv, X.-L.; Feng, D.; Xie, L.-H.; Zhang, J.; Li, M.; Xie, Y.; Li, J.-R.; Zhou, H.-C., Highly stable Zr (IV)-based metal–organic frameworks for the detection and removal of antibiotics and organic explosives in water. *J. Am. Chem. Soc.* **2016**, *138* (19), 6204-6216.
(37) Zheng, Z.; Rong, Z.; Iu‐Fan Chen, O.; Yaghi, O. M., Metal‐Organic Frameworks with Rod Yttrium Secondary Building Units. *Isr. J. Chem.* **2023**, e202300017.
(38) Song, W.; Zheng, Z.; Alawadhi, A. H.; Yaghi, O. M., MOF water harvester produces water from Death Valley desert air in ambient sunlight. *Nat. Water* **2023**, *1* (7), 626–634.
(39) Abrahams, B. F.; Dharma, A. D.; Grannas, M. J.; Hudson, T. A.; Maynard-Casely, H. E.; Oliver, G. R.; Robson, R.; White, K. F., Isomeric ionic lithium isonicotinate three-dimensional networks and single-crystal-to-single-crystal rearrangements generating microporous materials. *Inorg. Chem.* **2014**, *53* (10), 4956-4969.
(40) Zheng, Z.; Nguyen, H. L.; Hanikel, N.; Li, K. K.-Y.; Zhou, Z.; Ma, T.; Yaghi, O. M., High-yield, green and scalable methods for producing MOF-303 for water harvesting from desert air. *Nat. Protoc.* **2023**, *18*, 136–156.
(41) Park, K. S.; Ni, Z.; Côté, A. P.; Choi, J. Y.; Huang, R.; Uribe-Romo, F. J.; Chae, H. K.; O'Keeffe, M.; Yaghi, O. M., Exceptional chemical and thermal stability of zeolitic imidazolate frameworks. *Proc. Natl. Acad. Sci. U.S.A.* **2006**, *103* (27), 10186-10191.
(42) Zheng, Z.; Hanikel, N.; Lyu, H.; Yaghi, O. M., Broadly Tunable Atmospheric Water Harvesting in Multivariate Metal–Organic Frameworks. *J. Am. Chem. Soc.* **2022**, *144* (49), 22669-22675.
(43) Mitchell, M.; Palmarini, A. B.; Moskvichev, A. Comparing Humans, GPT-4, and GPT-4V On Abstraction and Reasoning Tasks. *arXiv preprint* DOI: 10.48550/arXiv.2311.09247 (accessed 2023-11-14).
(44) Rohatgi, A., WebPlotDigitizer. San Francisco, California, USA: 2019.
(45) Gagliardi, L.; Yaghi, O. M., Three Future Directions for Metal–Organic Frameworks. ACS Publications: 2023; Vol. 35, pp 5711-5712.

(46) AlGhamdi, A. A., Large Language Models, Computational Chemistry, and Digital Reticular Chemistry: A Perspective and Proposed Workflow. *Mol. Front. J.* **2023**, 1-4.

(47) Khattab, O.; Singhvi, A.; Maheshwari, P.; Zhang, Z.; Santhanam, K.; Vardhamanan, S.; Haq, S.; Sharma, A.; Joshi, T. T.; Moazam, H. Dspy: Compiling declarative language model calls into self-improving pipelines. *arXiv preprin* DOI: 10.48550/arXiv.2310.03714 (accessed 2023-10-05).

(48) Duan, X.; Wu, C.; Xiang, S.; Zhou, W.; Yildirim, T.; Cui, Y.; Yang, Y.; Chen, B.; Qian, G., Novel microporous metal–organic framework exhibiting high acetylene and methane storage capacities. *Inorg. Chem.* **2015**, *54* (9), 4377-4381.

(49) Furukawa, H.; Gandara, F.; Zhang, Y.-B.; Jiang, J.; Queen, W. L.; Hudson, M. R.; Yaghi, O. M., Water adsorption in porous metal–organic frameworks and related materials. *J. Am. Chem. Soc.* **2014**, *136* (11), 4369-4381.

(50) Marshall, R. J.; Hobday, C. L.; Murphie, C. F.; Griffin, S. L.; Morrison, C. A.; Moggach, S. A.; Forgan, R. S., Amino acids as highly efficient modulators for single crystals of zirconium and hafnium metal–organic frameworks. *J. Mater. Chem. A* **2016**, *4* (18), 6955-6963.

(51) Krap, C. P.; Newby, R.; Dhakshinamoorthy, A.; García, H.; Cebula, I.; Easun, T. L.; Savage, M.; Eyley, J. E.; Gao, S.; Blake, A. J., Enhancement of CO2 adsorption and catalytic properties by Fe-doping of [Ga2 (OH) 2 (L)](H4L= biphenyl-3,3′,5,5′-tetracarboxylic acid), MFM-300 (Ga2). *Inorg. Chem.* **2016**, *55* (3), 1076-1088.

(52) Volkringer, C.; Popov, D.; Loiseau, T.; Férey, G.; Burghammer, M.; Riekel, C.; Haouas, M.; Taulelle, F., Synthesis, single-crystal X-ray microdiffraction, and NMR characterizations of the giant pore metal-organic framework aluminum trimesate MIL-100. *Chem. Mater.* **2009**, *21* (24), 5695-5697.

(53) Liu, B.; Zhou, H.-F.; Hou, L.; Wang, J.-P.; Wang, Y.-Y.; Zhu, Z., Structural diversity of cadmium (II) coordination polymers induced by tuning the coordination sites of isomeric ligands. *Inorg. Chem.* **2016**, *55* (17), 8871-8880.

(54) Li, T.; Kozlowski, M. T.; Doud, E. A.; Blakely, M. N.; Rosi, N. L., Stepwise ligand exchange for the preparation of a family of mesoporous MOFs. *J. Am. Chem. Soc.* **2013**, *135* (32), 11688-11691.

(55) Yu, Q.; Zeng, Y.-F.; Zhao, J.-P.; Yang, Q.; Hu, B.-W.; Chang, Z.; Bu, X.-H., Three-dimensional porous metal− organic frameworks exhibiting metamagnetic behaviors: synthesis, structure, adsorption, and magnetic properties. *Inorg. Chem.* **2010**, *49* (9), 4301-4306.

# Supporting Information

# Image and Data Mining in Reticular Chemistry Using GPT-4V

Zhiling Zheng,[†,‡,§] Zhiguo He,[†,‡] Omar Khattab,[¶] Nakul Rampal,[†,‡,§] Matei A. Zaharia,[◊] Christian Borgs,[§,◊] Jennifer T. Chayes,[§,◊,○,∆] and Omar M. Yaghi[†,‡,§,∥,]*

† Department of Chemistry, University of California, Berkeley, California 94720, United States

‡ Kavli Energy Nanoscience Institute, University of California, Berkeley, California 94720, United States

§ Bakar Institute of Digital Materials for the Planet, University of California, Berkeley, California 94720, United States

¶ Department of Computer Science, Stanford University, Stanford, California 94305, United States

◊ Department of Electrical Engineering and Computer Sciences, University of California, Berkeley, California 94720, United States

○ Department of Mathematics, University of California, Berkeley, California 94720, United States

∆ Department of Statistics, University of California, Berkeley, California 94720, United States

⊥ School of Information, University of California, Berkeley, California 94720, United States

∥ KACST–UC Berkeley Center of Excellence for Nanomaterials for Clean Energy Applications, King Abdulaziz City for Science and Technology, Riyadh 11442, Saudi Arabia

* To whom correspondence should be addressed: yaghi@berkeley.edu

**Prompt**: Describe the image.

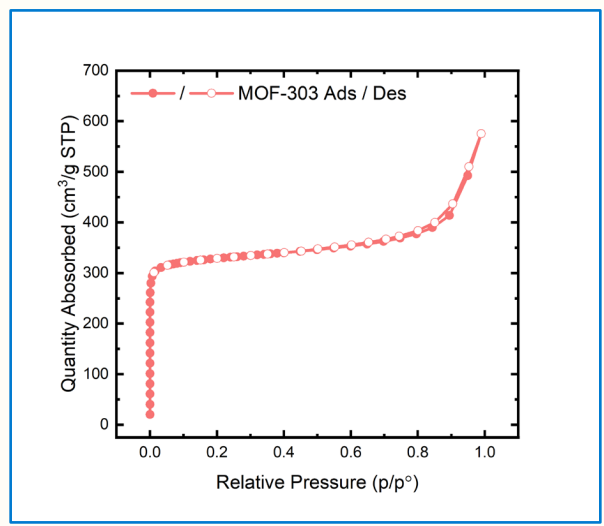

**GPT-4V**:
The image is a graph representing the adsorption and desorption isotherms of a material named MOF-303. On the x-axis, we have "Relative Pressure (p/p°)" which is dimensionless, and it ranges from 0 to 1. The y-axis represents "Quantity Absorbed (cm³/g STP)" which is a volume per mass at standard temperature and pressure (STP), and it ranges from 0 to 700 cm³/g.

The adsorption part of the isotherm (as pressure increases) appears to have a steep rise at low pressures indicating a rapid uptake of the gas. After a certain point, the graph plateaus, suggesting that the material has reached its maximum adsorption capacity.

The markers are open circles, connected by a line, which is common in such types of graphs to represent experimental data points. The color of the data points and line is a shade of pink or light red. The background is white, and the axes are labeled in black with a standard font.

**Figure S1.** Interaction with GPT-4V using a nitrogen sorption isotherm image. The provided image was combined with a text prompt to evaluate GPT-4V's proficiency in interpreting scientific figures from literature. Note that the data depicted are a replot from a previous publication[1] to avoid copyright concerns. In its response, GPT-4V demonstrated the ability to discern not only the content but also to interpret the colors, characters, and numbers within the graph.

**Prompt**: Describe the image.

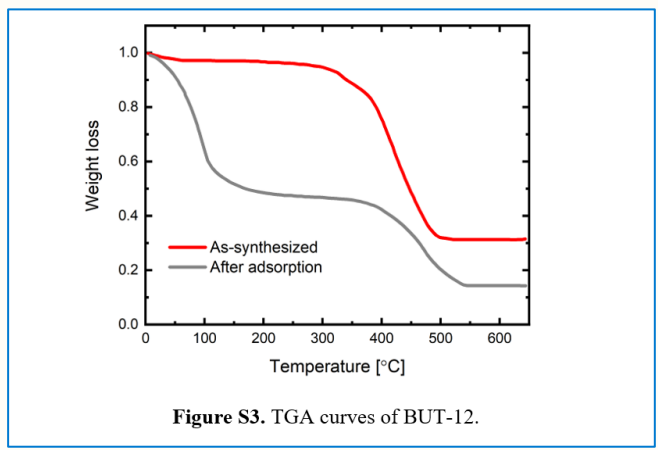

**Figure S3.** TGA curves of BUT-12.

---

**GPT-4V**:

The image is a graph, specifically a Thermogravimetric Analysis (TGA) curve. TGA is a type of testing that is used to determine changes in weight in relation to change in temperature. The graph shows two curves:

1. The red curve represents the "As-synthesized" sample, which indicates the weight loss of the material as it was originally produced.
2. The grey curve represents the sample "After adsorption", showing the weight loss after the material has presumably adsorbed some substance.

The x-axis of the graph indicates the temperature in degrees Celsius (°C), ranging from 0 to 600°C. The y-axis represents the weight loss, scaled from 0 to 1, which is likely a normalized or relative scale rather than absolute weight.

The red curve starts at a weight loss value close to 1 and remains stable until approximately 200°C, after which it begins to decline, indicating that the material is losing weight as the temperature increases, likely due to the loss of volatile components or decomposition. The grey curve starts also close to 1 but begins to decline slightly earlier than the red curve, and continues to do so at a gradual rate, suggesting that the adsorbed material affects the weight loss pattern.

At the bottom, there's a caption "Figure S3. TGA curves of BUT-12," which identifies this as the third supplemental figure in a series, related to a material or compound referred to as BUT-12. This kind of graph is commonly found in scientific publications, particularly in materials science and chemistry to report the thermal stability of materials.

**Figure S2.** Interaction with GPT-4V using a plot containing TGA curves. The provided image was combined with a text prompt to evaluate GPT-4V's proficiency in interpreting scientific figures from literature. Note that the data depicted are a replot from a previous publication[2] to avoid copyright concerns. In its response, GPT-4V demonstrated the ability to discern not only the content but also to interpret the colors, characters, and numbers within the graph.

**Prompt**: Describe the image.

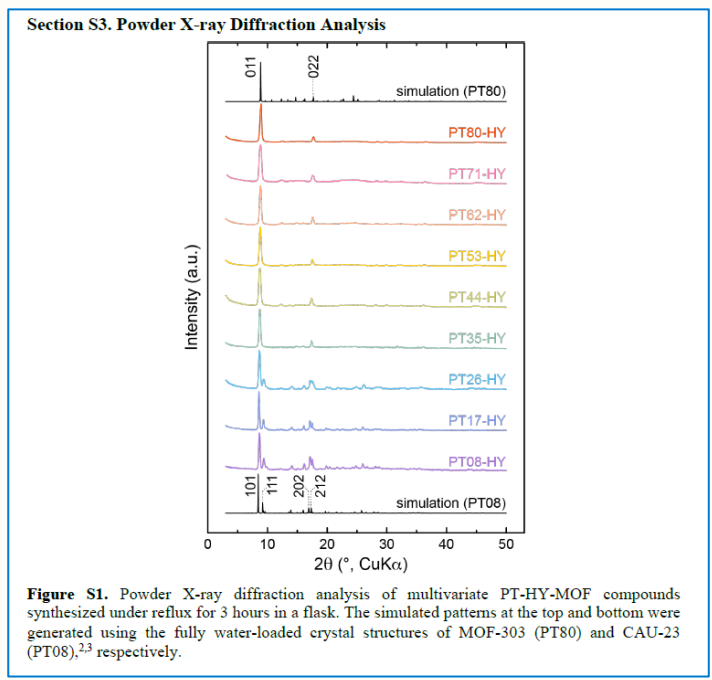

**GPT-4V**:

The image is of a scientific figure from a research paper or report, specifically Section S3 titled "Powder X-ray Diffraction Analysis". It shows a series of X-ray diffraction (XRD) patterns for different samples, which are labeled PT80-HY, PT71-HY, PT62-HY, PT53-HY, PT44-HY, PT35-HY, PT26-HY, PT17-HY, and PT08-HY. Each pattern displays the intensity of diffracted X-rays as a function of the angle (2θ, with CuKα radiation), which is a common method for characterizing the crystalline structures of materials.

At the top and bottom of the graph, there are simulated XRD patterns labeled as "simulation (PT80)" and "simulation (PT08)", which serve as references for comparing the experimental data. These simulated patterns are for the fully water-loaded crystal structures of two different materials referred to as MOF-303 (PT80) and CAU-23 (PT08), respectively. The intensity is plotted on the y-axis (in arbitrary units, a.u.), and the angle 2θ is on the x-axis, ranging from 0 to 50 degrees. The peaks in each pattern correspond to the crystalline planes within the materials, and their positions and intensities are unique to the specific structure of the material being analyzed.

The text below the figure describes the conditions under which the PT-HY-MOF compounds were synthesized (under reflux for 3 hours in a flask) and mentions the simulated patterns that correspond to the water-loaded crystal structures of MOF-303 (PT80) and CAU-23 (PT08). The footnote numbers "2,3" after "PT08" refer to citations within the main text of the document.

**Figure S3.** Interaction with GPT-4V using a PXRD pattern sourced from a previous publication[3]. Copyright © 2022 American Chemical Society.

**Prompt**: Describe the images.

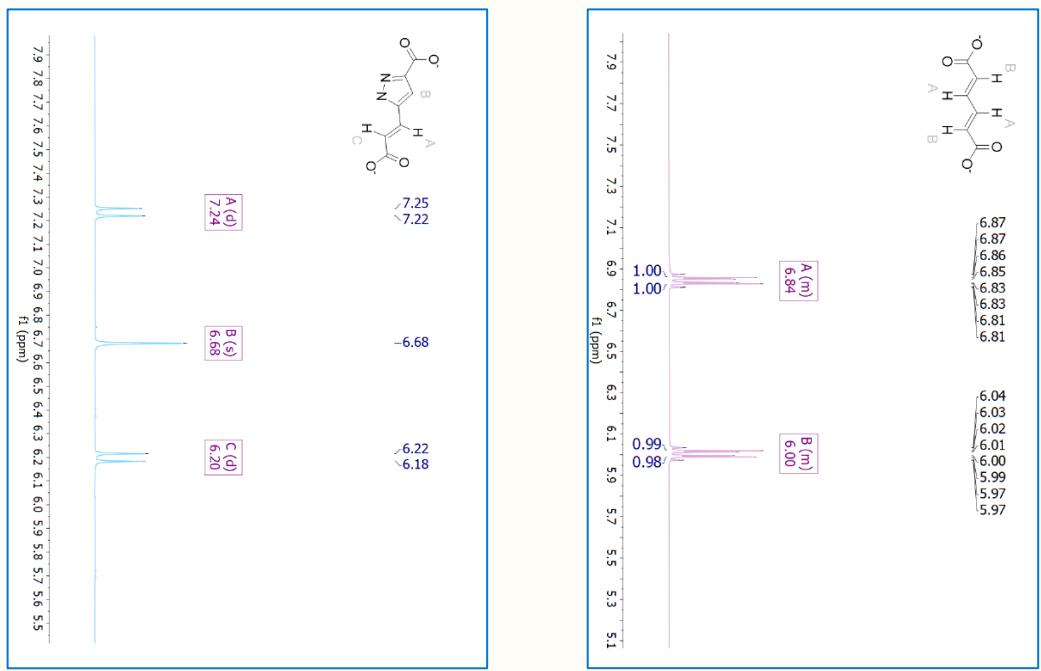

**GPT-4V**:
The images you've provided appear to be screenshots of nuclear magnetic resonance (NMR) spectra, which are used in chemistry to determine the structure of molecules. Each peak or set of peaks in the spectrum corresponds to particular groups of atoms within the molecule, and their position (chemical shift) on the horizontal axis (typically measured in parts per million, or ppm) provides information about the electronic environment of these atoms.

The accompanying chemical structures suggest that these spectra are being used to analyze these specific molecules, with annotations indicating the peaks corresponding to particular atoms or groups within the molecules. The letters (A, B, C, etc.) next to the peaks likely correlate to the labeled atoms or groups in the molecular structure shown in the image.

Such data is often used by chemists for compound verification, structure elucidation, and in studies of molecular dynamics. Without specific details on the molecules or the context of the spectra, this is a general interpretation of the type of information that these images represent.

**Figure S4.** Interaction with GPT-4V using two NMR spectra images rotated 90 degrees. GPT-4V demonstrated the ability to process multiple images simultaneously and identify the plots even when rotated.

**Prompt**: Describe the image.

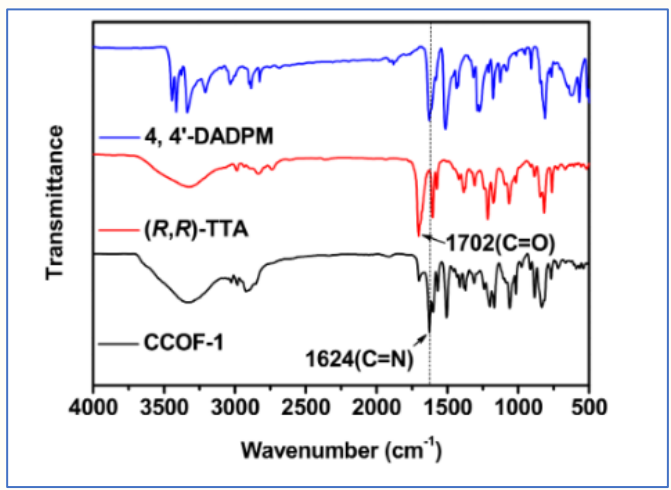

**GPT-4V**:
The image shows an infrared (IR) spectroscopy graph with three overlaid spectra, each represented by a different color. From top to bottom, there is a blue, a red, and a black spectrum.

- The blue spectrum is labeled as "4, 4'-DADPM".
- The red spectrum is labeled as "(R,R)-TTA".
- The black spectrum is labeled as "CCOF-1".

There are two peaks in the spectra with annotations:
- One peak in the red spectrum is labeled with "1702 (C=O)", indicating the transmittance peak corresponding to a carbonyl functional group.
- A peak in the black spectrum is labeled with "1624 (C=N)", indicating a transmittance peak for a carbon-nitrogen double bond.

The x-axis is labeled as "Wavenumber (cm^-1)", which is the standard unit of wavenumber for IR spectra, ranging from approximately 4000 to 500 cm^-1. The y-axis is labeled as "Transmittance", which measures how much light is transmitted at each wavenumber. The transmittance is plotted as a percentage, where a higher transmittance indicates less absorption by the sample at that particular wavenumber.

This type of graph is typically used in analytical chemistry to identify functional groups in molecules and to characterize chemical compounds. Each of the labeled peaks corresponds to a different molecular vibration that is characteristic of a certain chemical bond or functional group within the molecule.

**Figure S5.** Interaction with GPT-4V using the IR spectra sourced from a previous publication[4]. Copyright © 2016 American Chemical Society.

**Prompt**: Describe the image.

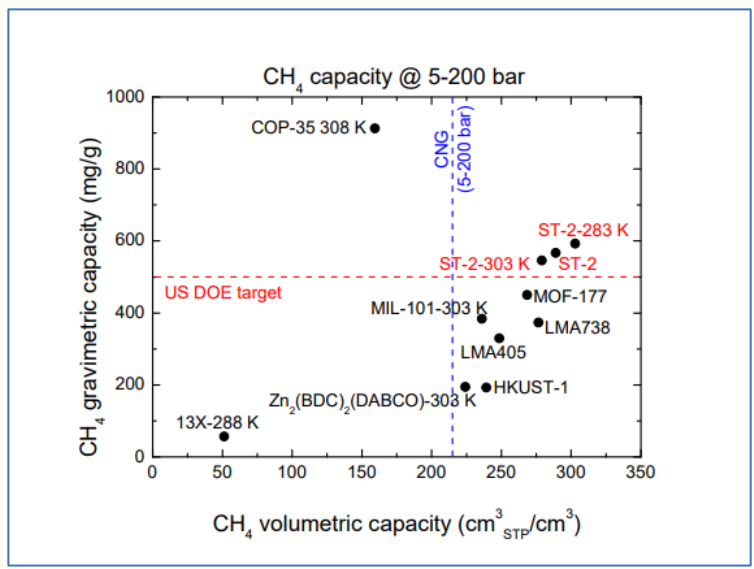

**GPT-4V**:
The image is a graph plotting methane (CH4) storage capacities of various materials under a pressure range of 5-200 bars. The x-axis represents the volumetric capacity (measured in cubic centimeters of standard temperature and pressure gas per cubic centimeter of material, cm³_STP/cm³), while the y-axis shows the gravimetric capacity (measured in milligrams of gas per gram of material, mg/g).

Different materials are plotted as points on the graph, such as COP-35, Zn2(BDC)2(DABCO), MIL-101, MOF-177, HKUST-1, and others, each represented by a distinct marker. The U.S. Department of Energy (DOE) target for methane storage is indicated by a dashed horizontal red line, while a dashed vertical blue line may indicate a benchmark or target for volumetric capacity.

The graph is used to compare the performance of these materials against the DOE targets for methane storage, which is a key metric for applications like natural gas vehicles where storage efficiency can determine the range and practicality of the technology.

**Figure S6.** Interaction with GPT-4V using a scatter plot sourced from a previous publication[5]. GPT-4V demonstrated the capability to identify key elements within the plot and to interpret them accurately. Copyright © 2017 American Chemical Society.

**Prompt**: Describe the image.

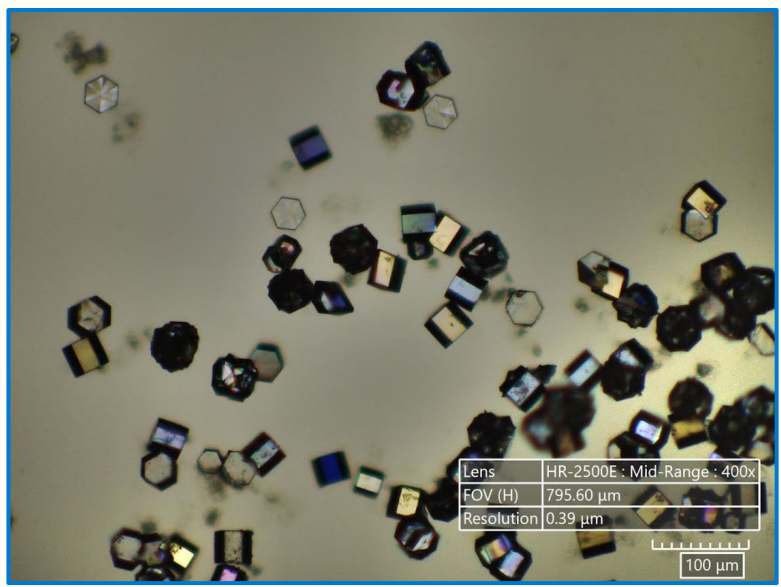

**GPT-4V**:

This image appears to be taken through a microscope, showing a variety of crystal-like structures, possibly a salt or some mineral crystalline forms, scattered across the field of view. The crystals have different shapes, with some exhibiting more complex geometries, and many of them display dark edges, likely due to their refractive properties bending light. The colors vary from clear to dark shades, possibly due to the thickness or orientation of the crystals.

On the bottom of the image, there's a scale bar indicating 100 micrometers (μm), which provides a reference for the size of the crystals. The information overlaid on the image indicates that this is taken with a high-resolution lens (HR-2500E) at a mid-range magnification of 400x. The field of view (FOV) is noted as 795.60 micrometers horizontally, with a resolution of 0.39 micrometers, suggesting that the image is highly detailed.

**Figure S7.** Interaction with GPT-4V using a microscopic image. The image was accompanied by a textual prompt to assess GPT-4V's ability to interpret real experimental result images, including labels and scale bars. Notably, the image is from unpublished work, ensuring that GPT-4V's interpretation is based on an image it has not previously encountered, thus testing its ability to recognize new crystal images.

**Prompt**: Describe the image.

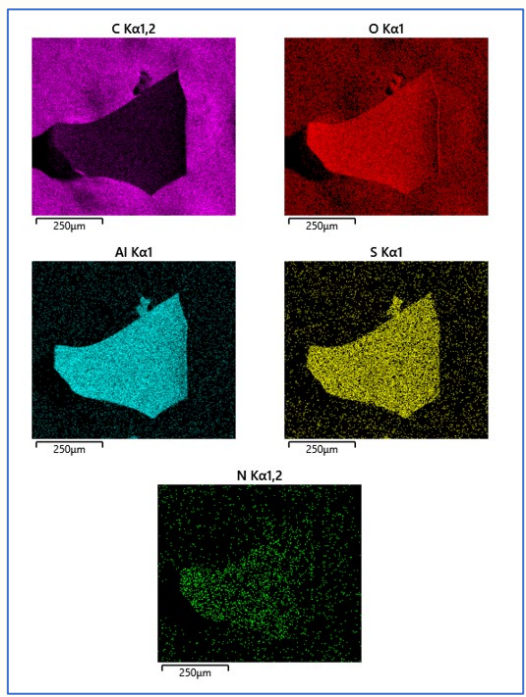

**GPT-4V**:
The image appears to be a composite of five separate images, each one showing a different Energy-dispersive X-ray spectroscopy (EDX or EDS) map of a sample. These maps are used to illustrate the distribution of elements within the sample. Each of the images is labeled with the element it represents and the characteristic X-ray line used in the analysis (Kα line). The elements displayed are carbon (C), oxygen (O), aluminum (Al), sulfur (S), and nitrogen (N).

Each map is color-coded for a specific element, and the intensity of the color represents the concentration of the element in the sample. The colors used are purple for carbon, red for oxygen, cyan for aluminum, yellow for sulfur, and green for nitrogen. Each image also has a scale indicator in the lower left corner, stating "250µm," which implies that the field of view is 250 micrometers wide.

**Figure S8.** Interaction with GPT-4V using a series of scanning electron microscopy (SEM) with energy dispersive X-ray spectroscopy images. GPT-4V was able to identify and interpret the provided image. Note that the images are from unpublished work, ensuring that GPT-4V's interpretation is based on an image it has not previously encountered, thus testing its ability to recognize SEM images.

**Prompt**: Describe the image.

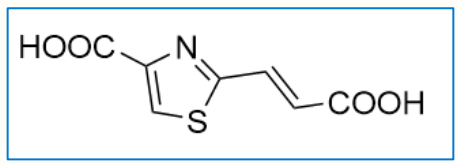

**GPT-4V**:
The image you've uploaded appears to be a structural chemical formula. It represents an organic molecule with a specific arrangement of atoms and bonds. The molecule contains a ring structure that includes nitrogen (N) and sulfur (S) atoms, indicating it might be a heterocyclic compound. There are two carboxylic acid groups (-COOH) attached to the ring, suggesting the compound is a dicarboxylic acid. The precise angles and lengths of the lines in the diagram depict the bonding structure and the molecular geometry of the compound. This type of diagram is commonly used in chemistry to represent the structure of molecules.

**Figure S9.** Interaction with GPT-4V using a chemical structure of an organic linker that has never been published in any paper or dataset before.

**Prompt**: Describe the image.

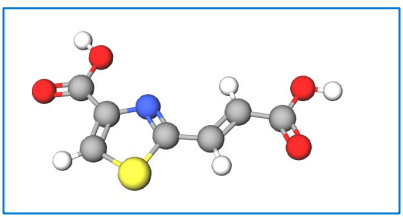

**GPT-4V**:
The image appears to be a three-dimensional molecular structure model, which is commonly used in chemistry to represent the arrangement of atoms in a molecule. This particular model shows a molecule with various colored spheres representing different atoms. The grey spheres typically represent carbon atoms, the white spheres are hydrogen, the red spheres are oxygen, and the blue sphere is likely nitrogen. The yellow sphere could represent a sulfur atom. These spheres are connected by sticks which represent chemical bonds; single bonds are typically depicted by a single stick, and double bonds by two sticks parallel to each other.

This kind of representation is known as a ball-and-stick model and is used to provide a visual interpretation of a molecule's structure, showing how the atoms are connected and their relative positions in three-dimensional space.

**Figure S10.** Interaction with GPT-4V using a 3D version of the same molecule with different chemical elements in various color settings.

**Prompt**: Describe the image.

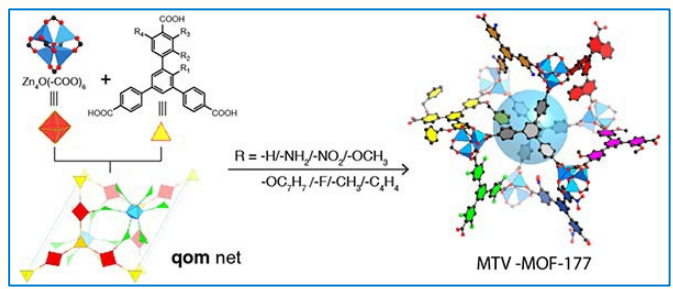

**GPT-4V**: The image is a graphical representation of chemical processes and structures. On the left side, there is a reaction scheme showing a metal ion, Zn(II), reacting with a ligand that is a complex organic molecule with a central benzene ring substituted with various functional groups and carboxylic acid groups. The ligand is depicted with R groups, indicating variable substituents, which include -H, -NH₂, -NO₂, -OCH₃, -OC₆H₄, -F, and -CH₃-C₆H₄. Below this reaction scheme, there is a colorful illustration of a "qom net," which seems to represent a type of molecular framework or structure resulting from this reaction.

On the right side, there is a more complex molecular structure labeled "MTV-MOF-177," which is likely a type of Metal-Organic Framework (MOF). This part of the image shows a 3D ball-and-stick model with various colors representing different atoms or groups within the structure. The central part of the structure is highlighted with a semi-transparent blue sphere, indicating a cavity or pore within the MOF, which is typical for these materials known for their porosity.

The image is likely used in a scientific context, possibly in materials science or chemistry, to illustrate the synthesis of a Metal-Organic Framework and the resulting porous structure that can have applications in gas storage, separation, or catalysis.

**Prompt**: What does MTV mean here?

**GPT-4V**: In the context of Metal-Organic Frameworks (MOFs), MTV stands for "multivariate," which refers to the incorporation of multiple organic functionalities on the ligands within a single framework. In MTV-MOFs, different functionalities occupy the same kind of positions in the framework, and the percentage of each functionality can be varied.

**Figure S11.** Interaction with GPT-4V using a TOC image from a publication abstract, which details the schematic synthesis of a MOF described in the publication[6]. GPT-4V's response accurately identifies key components: the metal ion, the organic linker, and the resulting MOF structure and topology. Additionally, the dialogue demonstrates GPT-4V's capability to delve deeper into specific aspects, such as explaining the abbreviation 'MTV' in the context of MOFs. Copyright © 2015 American Chemical Society.

**Prompt**: Describe the image.

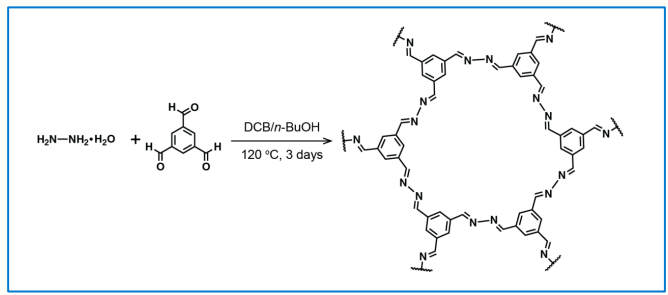

**GPT-4V**: The image displays a chemical synthesis scheme labeled "Scheme S3. Synthesis of AB-COF." It illustrates a chemical reaction where two molecules on the left side are reacting to form a more complex molecule on the right side. On the left, there are two reactants:

1. A molecule of hydrazine hydrate ($H_2N-NH_2*H_2O$).
2. A molecule that appears to be a derivative of terephthalaldehyde.

These reactants are subjected to a reaction condition using a catalyst or reagent (DCB/n-BuOH) at a temperature of 120°C for 3 days. The product on the right is a large, ring-shaped organic molecule, which likely represents a covalent organic framework (COF), given the label "AB-COF." This molecule is composed of repeating units that are connected by nitrogen atoms, forming a hexagonal mesh-like structure. The repeating units include aromatic rings with nitrogen and carbon atoms, indicating a possible conjugated system which could contribute to the properties of the AB-COF.

The specific arrangement of the atoms and the bonds suggests that the COF has been synthesized through a condensation reaction, where smaller molecules join together, releasing smaller by-products such as water, and creating a larger, more complex structure. The presence of nitrogen in the hexagonal rings indicates that the reaction involves the formation of azine or imine bonds.

**Prompt**: Tell me more about the reaction condition.

**GPT-4V**: The reaction condition "DCB/n-BuOH at 120°C, 3 days" indicates that the synthesis of the AB-COF is performed in a mixed solvent system consisting of dichlorobenzene (DCB) and n-butanol (n-BuOH) at a high temperature of 120 degrees Celsius, maintained for three days. DCB is a non-polar solvent, while n-BuOH is a moderately polar solvent; their combination may be used to optimize the solubility of the reactants and control the reaction rate. The high temperature and extended time suggest a thermally driven reaction that may require such conditions for the reactants to fully convert into the covalent organic framework (COF).

**Figure S12.** Interaction with GPT-4V using a synthesis scheme from previous publication[7]. GPT-4V's response identifies the starting materials, reaction conditions and products. In addition, it is able to further explain chemicals' abbreviations. Copyright © 2019 American Chemical Society.

**Prompt**: Describe the image.

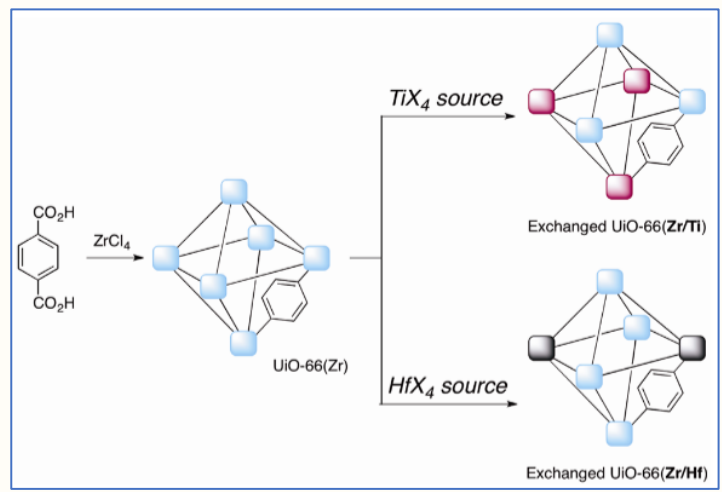

**GPT-4V**:
The image is a schematic representation of a chemical process, specifically the synthesis and post-synthetic modification of a metal-organic framework (MOF) known as UiO-66.

On the left, there is a chemical structure with two carboxylic acid groups ($CO_2H$) attached to a benzene ring, which is the organic linker used to synthesize the MOF. This linker reacts with $ZrCl_4$ (zirconium tetrachloride), represented by blue spheres, to form the UiO-66(Zr) framework. The blue spheres likely represent $Zr^{4+}$ ions, and the lines between them and the organic linker represent coordination bonds.

On the right side, there are two pathways shown for post-synthetic modification of UiO-66(Zr). The upper pathway shows the MOF reacting with $TiX_4$ (where X represents halogen atoms), resulting in an "Exchanged UiO-66(Zr/Ti)" where some $Zr^{4+}$ ions in the framework have been replaced by $Ti^{4+}$ ions, indicated by pink spheres.

The lower pathway shows the MOF reacting with $HfX_4$ (where X again represents halogen atoms), resulting in "Exchanged UiO-66(Zr/Hf)" where some $Zr^{4+}$ ions have been replaced by $Hf^{4+}$ ions.

The image is used to illustrate the synthetic pathway of UiO-66 and how its metal sites can be modified post-synthesis with other metal sources like Ti and Hf. The representations are stylized rather than realistic, with colors and shapes indicating different components of the MOF structure.

**Figure S13.** Interaction with GPT-4V using a post synthetic cation exchange scheme from previous publication[8]. GPT-4V's response clearly explains the process, starting from the synthesis to create the pristine MOF and detailing how it was exchanged. Copyright © 2012 American Chemical Society.

**Prompt**: Describe the image.

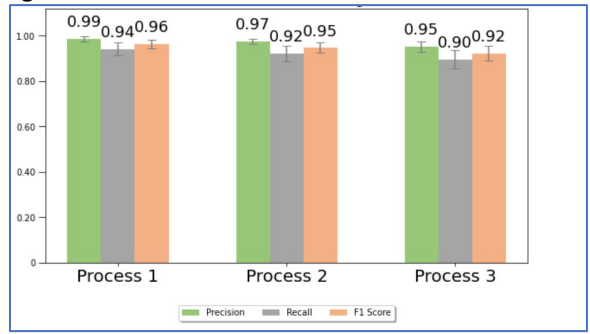

---

**GPT-4V**: The image is a bar chart with three groups of bars, each group representing a different process: Process 1, Process 2, and Process 3. For each process, there are three bars in different colors, each color representing a metric: Precision (green), Recall (orange), and F1 Score (gray).

- For Process 1, Precision is approximately 0.99, Recall is about 0.94, and the F1 Score is around 0.96.
- In Process 2, Precision drops slightly to around 0.97, Recall is roughly 0.92, and the F1 Score is about 0.95.
- For Process 3, the Precision is around 0.95, Recall is approximately 0.90, and the F1 Score is close to 0.92.

Each bar has a black line at the top, representing the error bar, which indicates the variability or uncertainty in the measurements of each metric. The scale on the y-axis goes from 0 to 1 in increments of 0.2, and each bar reaches up to a value between 0.9 and 1.0, indicating high performance across all three metrics for all processes.

---

**Prompt**: Write python codes to plot a similar image.

---

**GPT-4V**:

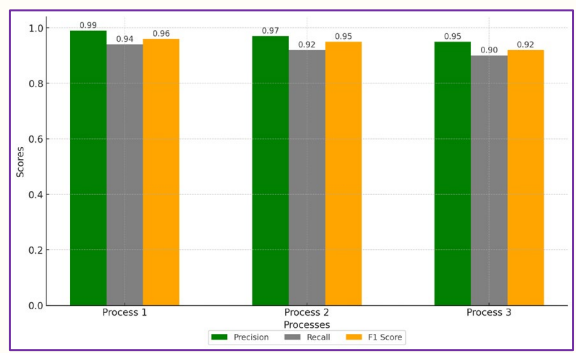

**Figure S14.** Interaction with GPT-4V using a performance metrics plot sourced from a previous publication[9]. Further requests can be made to replot the image based on the information seen by GPT-4V in the provided image.

Please analyze the provided image, which is a page from a literature paper discussing metal-organic frameworks, and answer the following questions:

1.Does the image contain one of the following figures:
 (1) Nitrogen adsorption-desorption isotherm
(2) PXRD pattern
(3) TGA curve
(4) Crystal structure or topology illustration
(5) Other gas sorption isotherm (e.g. water vapor, co2, H2, CH4, etc.)
(6) It does not have figures or has figures that do not belong to the above class

2. If the figure contains a nitrogen adsorption-desorption isotherm, in which figure (e.g. Figure 1, 2, 3, Figure S18) or panel (e.g., a, b, c, d, if applicable) is it displayed? Note that we are interested only in the nitrogen adsorption-desorption isotherms that show the direct relationship between pressure and nitrogen uptake, explicitly excluding BET analysis, linear fitting, pore size distribution graphs, and any other analyses that do not present this direct relationship and do not label these cases as (1).
3. If a nitrogen isotherm is present, can you identify the name of the compound being measured? If there are multiple compounds in the same nitrogen isotherm, please separate your answer by a comma. Please note that for figure labeling, if there are adsorption (ads) and desorption (des) plots for the same compound, the author tends to label them as ads xxx or xxx adsorption or des xxx or xxx desorption, you should treat them as one compound and list the name xxx only once, usually a compound name does not include ads or des.
4. Does the page provide information on porosity information such as surface area or nitrogen uptake for the compound? If not provided on this page, please indicate with "N/A."
5. For each nitrogen isotherm, identify the saturation amount of nitrogen gas in the plateau region of the adsorption branch. The plateau region refers to the area where the curve levels off before any significant further increase in gas uptake, even as the relative pressure (P/P0) continues to rise. The saturation level should be determined from the range within this plateau, ignoring extreme points or outliers (e.g. those at the top right corner where P/P0> 0.9). Report the value in a range of the y values given in the plot next to the y-axis which can be read (e.g., 300 - 400 cm3/g). If you are unsure about your answer, please indicate with "I do not know."
6. Localize the region in this page containing nitrogen isotherm plot and its Figure caption and description of nitrogen isotherm plot. Regions are represented by (x1,y1,x2,y2) coordinates. x1 and x2 are the leftmost and rightmost positions, normalized into 0 to 1, where 0 is the left and 1 is the right. y1 and y2 are the topmost and bottommost positions, normalized into 0 to 1, where 0 is the top and 1 is the bottom. For example, if Figures 5(a) and 5(b) are both nitrogen isotherm and figure 5(c) is something else, you tell me the box that cover top left corner coordinate and bottom right coordinate of entire figure 5 including caption in the format like (0,0,1,0.5). Please try to make your selection to include all elements, especially that in that figure and its caption and it is okay if you use larger box selection which results in the inclusion of some parts like text or blank space or journal logo or something else that is outside figure 5. On the other hand, if Figure 5 and Figure 7(c) are nitrogen isotherms, while Figure 6 and Figures 7(a) and 7(b) are something else, you will give two regions, one specify the entire Figure 5, and the other one cover all elements in Figure 7.

Please strictly adhere to the templates provided.
***
Template for your answer if no nitrogen adsorption-desorption isotherm is found:

Figures: [Answer: (2) ... (6) or two or more combinations of them like (2);(3)]
Nitrogen Isotherm: No

Template for your answer if a nitrogen isotherm is presented:
Figures: [Answer: (1), (2) ... (5) or two or more combinations of them like (1);(4);(5)]
Nitrogen Isotherm: [Answer: specify figure and/or panel]
Compound: [Answer: Compound name or "N/A"]
Porosity: [Answer: Specific value/details or "N/A"]
Hysteresis: [Answer: Yes or No or "I do not know"]
Saturation: [Answer: Specific range or "I do not know"]
Position: [Answer: Specify region(s) (x 1, y 1, x 2, y 2)]
***

**Figure S15.** Full prompt guiding GPT-4 for literature figure analysis in this study.

```python
import fitz, os, csv

def pdf_to_images(pdf_path, output_folder, dpi=300):
    # Open PDF and set up image conversion settings
    doc, scale_factor = fitz.open(pdf_path), dpi / 72.0
    if not os.path.exists(output_folder): os.makedirs(output_folder)

    # Convert each page to image and collect data
    image_data = [(f"{os.path.basename(pdf_path).replace('.pdf', '')}_page_{page_num+1}.png",
page_num+1)
                  for page_num, page in enumerate(doc) if
(page.get_pixmap(matrix=fitz.Matrix(scale_factor,
scale_factor)).save(os.path.join(output_folder, f"{os.path.basename(pdf_path).replace('.pdf',
'')}_page_{page_num+1}.png")))]
    return image_data

base_folder, folders_to_process = base, ["10.1016", "10.1021", "10.1038", "10.1039",
"10.1107"]
# Generate unique CSV file name
csv_path = next(os.path.join(os.getcwd(), f"dictionary_{counter}.csv") for counter in
range(1000) if not os.path.exists(os.path.join(os.getcwd(), f"dictionary_{counter}.csv")))
csv_data = []

# Process each folder and PDF, and collect image data for CSV
for foldername in folders_to_process:
    folder_path = os.path.join(base_folder, foldername)
    if os.path.isdir(folder_path):
        csv_data += [(f"{foldername}/{filename.replace('-si.pdf', '').replace('.pdf', '')}",
foldername, filename, image_name, page_num)
                     for filename in os.listdir(folder_path) if filename.endswith(".pdf")
                     for image_name, page_num in pdf_to_images(os.path.join(folder_path,
filename), folder_path)]

# Write data to CSV
with open(csv_path, 'w', newline='') as csvfile:
    csv.writer(csvfile).writerows([["DOI", "Folder Name", "PDF Name", "Image Name", "Page
Number"]] + csv_data)
```

**Figure S16.** Python code to convert PDF literature files into PNG image files by page, with each PDF file divided into multiple images, one image per page.

```python
import pandas as pd
import requests
import base64

def encode_image(image_path): # Function to encode the image for OpenAI API
    with open(image_path, "rb") as image_file:
        return base64.b64encode(image_file.read()).decode('utf-8')

def analyze_image(image_path, api_key): # Function to analyze the image using OpenAI's API
    prompt = """..."""  # (Place prompt here)
    base64_image = encode_image(image_path)
    headers = {"Content-Type": "application/json", "Authorization": f"Bearer {api_key}"}
    payload = {
        "model": "gpt-4-vision-preview",
        "messages": [{"role": "user", "content": [{"type": "text", "text": prompt}, {"type": "image_url", "image_url": {"url": f"data:image/jpeg;base64,{base64_image}"}}]}],
        "max_tokens": 300
    }
    try:
        response = requests.post("https://api.openai.com/v1/chat/completions", headers=headers, json=payload)
        response.raise_for_status()
        return response.json()['choices'][0]['message']['content']
    except Exception as e:
        print(e)
        return ""

def main_process(api_key, csv_file_path): # Main processing function
    img_api_df = pd.read_csv(csv_file_path)
    for index, row in img_api_df.iterrows():
        img_path = row['img']
        if pd.isna(row['text']) or "Error:" in str(row['text']).strip():
            print(f"Processing Image No. {index+1}: {img_path}")
            text = analyze_image(img_path, api_key)
            if text:
                img_api_df.at[index, 'text'] = text
                img_api_df.to_csv(csv_file_path, index=False)
            else:
                print(f"Failed to analyze image: {img_path}")
        else:
            print(f"Image No. {index+1} already processed.")
    print("Processing completed!")

if __name__ == '__main__':
    api_key = "your-api-key"  # Replace with your actual API key
    csv_file_path = 'img_API.csv'  # Path to your CSV file
    main_process(api_key, csv_file_path)
```

**Figure S17.** Python code to access GPT-4 via API. This script encodes images into a base64 format and sends them, along with a textual prompt, to the OpenAI GPT-4 API for analysis. The script iterates through a CSV file, processing images referenced by their file names.

```python
import pyautogui, pandas as pd, json, pyperclip, time, sys

# Define constants
TimeSleep, exit_flag, PointFileName= 60, False, "./config/points.json"

# Function to get region coordinates
def get_region(screen_width, screen_height, region_name):
    region_width, region_height = screen_width // 2, screen_height // 2
    regions = {'left': (0, 0, region_width, screen_height), 'right': (region_width, 0,
region_width, screen_height),
               'down': (0, region_height, screen_width, region_height)}
    if region_name in regions: return regions[region_name]
    raise ValueError("Invalid region name.")

# Function to find 'Enter' button on screen
def findEnter():
    match = pyautogui.locateOnScreen('./config/enter.png',
region=get_region(*pyautogui.size(), 4))
    return pyautogui.center(match) if match else None

# Function to process image and get result
def GetResult(imgPath):
    points, Recjson, error = json.load(open(PointFileName)), json.load(open(RecFileName)),
False
    pyautogui.click(points['create_dialogue']) # Create a new dialogue
    pyautogui.click(points['click_image_upload_button'], clicks=2) # Upload the image
    pyautogui.click(points['click_text_input_box'])
    pyperclip.copy(imgPath)
    pyautogui.hotkey('ctrl', 'v') # Paste image path to the input box for uploading
    pyautogui.press('enter')
    for _ in range(20):
        if exit_flag: sys.exit()
        time.sleep(1)
        if findEnter(): break
    else: return "Error: Upload failed.", True
    pyautogui.click(findEnter()[0] - 100, findEnter()[1]) # Find the input text box
    pyperclip.copy(question)
    pyautogui.hotkey('ctrl', 'v')
    pyautogui.press('enter')
    for _ in range(100): # Image uploading is successful when enter bottom turns grey
        if exit_flag or pyautogui.locateOnScreen('./config/grayEnter.png',
region=get_region(*pyautogui.size(), 4)): break
        pyautogui.click(points['blank_area']) # Click blank area on the right and scroll down
        pyautogui.scroll(-3000)
        time.sleep(1)
    pyautogui.click(points['blank_area']) # Scroll all the way down to the bottom
    pyautogui.scroll(-3000)
    for _ in range(20): # Detect the "Copy" icon next to the output text
        if exit_flag or pyautogui.locateOnScreen('./config/copy.png',
region=get_region(*pyautogui.size(), "right")): break
```

S18

```python
            pyautogui.click(points['blank_area'])
            pyautogui.scroll(-3000)
            time.sleep(0.5)
        else: return "Error: Output Not Found", True
    pyautogui.click(pyautogui.center(pyautogui.locateOnScreen('./config/copy.png', 
region=get_region(*pyautogui.size(), "right"))))
    return pyperclip.paste(), error

# Main execution block
if __name__ == '__main__':
    t = threading.Thread(target=check_keyboard_input)
    t.daemon = True
    t.start()
    df = pd.read_csv('img.csv', encoding="utf-8")
    ErrorTime = 0
    errorFlag = False
    for index, imgPath in enumerate(df['img']):
        if errorFlag == True:
            errorFlag = False
            time.sleep(TimeSleep)
            if ErrorTime == 3 :
                time.sleep(60*30) # Reach time limit, retry after 30 minutes
                ErrorTime = 0 #reset counting
        if (df.loc[index, "text"].strip() == "" or pd.isna(df.loc[index, 'text']) or 
len(df.loc[index, "text"].strip()) <= 5 or any(keyword in df.loc[index, 'text'] for keyword 
in ["Error:", "upload the image", "analyze", "Sorry", "sorry"]) or ("Figures" not in 
df.loc[index, 'text'] and "Nitrogen Isotherm" not in df.loc[index, 'text'])):
            text, errorFlag = GetResult(imgPath)
            df.loc[index, 'text'] = text
            df.to_csv("img.csv", encoding='utf-8', index=False)
            if errorFlag == True:
                ErrorTime +=1
                print(text)
                continue
            else:
                ErrorTime = 0 # reset error time
```

**Figure S18.** Python code to access GPT-4 via web interface. The script automates interactions with a web-based platform to retrieve GPT-4 responses. It utilizes 'config.png' files for graphical element recognition, guiding the script to locate specific buttons and input fields on the screen. 'Points' refer to pre-defined coordinates for dialogue creation, image upload buttons, and text input fields in the script, corresponding to these graphical elements, ensuring accurate navigation and interaction within the web interface.

```python
import pandas as pd
import re

def extract_choices(output):
    # Normalize the string to handle different cases and remove square brackets
    normalized_output = output.replace('\n', ' ').replace('[', '').replace(']', '').strip()

    # Check if 'Figures:' is in the output
    if 'Figures:' not in normalized_output:
        return [6]  # Default to choice 6 in these cases

    # Extract the part of the output before 'Nitrogen Isotherm:' if it's present
    nitrogen_isotherm_index = normalized_output.find('Nitrogen Isotherm:')
    if nitrogen_isotherm_index != -1:
        normalized_output = normalized_output[:nitrogen_isotherm_index]

    # Extract the part of the output starting from 'Figures:'
    figures_index = normalized_output.find('Figures:')
    if figures_index != -1:
        normalized_output = normalized_output[figures_index:]

    # Find all occurrences of choices in the format (number), making sure they are within 1-6
    choices_str_list = re.findall(r'\((\d)\)', normalized_output)
    choices = sorted(set([int(num) for num in choices_str_list if int(num) in range(1, 7)]))

    return choices

if __name__ == '__main__':

    # Read the CSV file
    df = pd.read_csv('img.csv')  # Read the CSV file
    # Apply the function to each row in 'text' column and store the result in a new column
    df['GPT-4V Output'] = df['text'].apply(extract_choices)
    # Write the updated DataFrame back to the CSV file
    df.to_csv('img.csv', index=False)
```

**Figure S19.** Python code to extract the label from GPT-4V's responses. This script processes textual output from GPT-4V, specifically targeting and extracting categorized choices. It normalizes the response text for consistent processing, removes extraneous characters, and identifies distinct numerical labels indicating different response types or classifications

```python
import pandas as pd
import re

def extract_nitrogen_info(row):
    output = row['text']
    info = {'Nitrogen Isotherm': "N/A", 'Compound': "N/A", 'Porosity': "N/A", 'Hysteresis': "N/A",
            'Saturation': "N/A", 'Position': "N/A"}

    # Extracting and cleaning each piece of information
    for key in info.keys():
        try:
            # Extracting the value
            value = output.split(f'{key}: ')[1].split('\\n')[0]

            # Removing the "[Answer XXXXX]" pattern
            clean_value = re.sub(r'\\[\\s*Answer\\s*:\\s*(.*?)\\s*\\]', r'\\1', value)

            # Further cleaning the value by removing unwanted characters
            clean_value = ''.join(filter(lambda char: char.isalnum() or char in " .,-;/()[]",
clean_value))

            # Updating the info dictionary
            info[key] = clean_value.strip() if clean_value else "N/A"
        except IndexError:
            continue  # If the key is not found, it remains "N/A"

    return pd.Series(info)

if __name__ == '__main__':

    # Load the CSV file
    file_path = 'img.csv'
    data = pd.read_csv(file_path)

    # Applying the extraction function to each row
    extracted_data = data.apply(extract_nitrogen_info, axis=1)

    # Updating the dataframe with the extracted data
    data.update(extracted_data)

    # Saving the updated dataframe to the CSV file
    data.to_csv(file_path, index=False)
```

**Figure S20.** Python code to extract the nitrogen isotherm information from GPT-4V's responses. This script eliminates irrelevant characters and extracts specific parameters from GPT-4V's formulated answers. These parameters include the figure number of the nitrogen isotherm, the name of the compound, the indicated porosity, the presence of hysteresis, the saturation plateau, and the position of the bounding box. The process ensures accurate and automated extraction of crucial information relevant to nitrogen isotherm from GPT-4V's responses.

**Table S1.** Comparative evaluation of accuracy in recognizing different plot types from published literature. This table presents a comparison of accuracy rates between web interface and API methods for recognizing various plot types. Note that the API method was tested on the first 1,000 images due to API request limitations, whereas the web interface results reflect an analysis of all 6,240 images.

| Plot Type | Web Accuracy | API Accuracy |
|---|---|---|
| Nitrogen Isotherm | 99.5% | 97.9% |
| Power X-Ray Diffraction | 99.2% | 95.3% |
| Thermogravimetric Analysis | 99.2% | 94.9% |
| Crystal Structure or Topology | 98.1% | 91.8% |
| Other Gas Sorption Isotherm | 95.0% | 96.9% |
| None of Above | 94.3% | 88.7% |

**Table S2.** Comparative analysis of image processing capabilities across 6240 pages from 347 peer-reviewed publications.

| Parameter | Web | API |
|---|---|---|
| Cost (USD) | $20 | $125 |
| Processing Time (days) | 12 | 0.4 |
| Number of Images | 6240 | 6240 |
| Number of Context Tokens | N/A | 122,221,798 |
| Number of Generated Tokens | 203,266 | 118,866 |
| Average Cost per Paper (USD) | $0.057 | $0.361 |

Note: This study analyzed a total of 347 papers, collectively encompassing 6,240 pages. For the Web Interface, a single account, priced at $20 per month, can process approximately 500 images/pages daily. The input context token is not applicable, as the token count for uploads via the web interface depends on the resolution and size of the figure. The count of generated tokens for the Web Interface is based on the outputs collected. For the API, each request costs approximately 2000 tokens, equating to $0.02 per image/page, with an average processing time of 5.2 seconds per request.

**Table S3.** Comparative analysis of image processing capabilities within a fixed budget of $100.

| Parameter | Web | API |
|---|---|---|
| Cost (USD) | $100 | $100 |
| Processing Time (days) | 30 | 0.3 |
| Estimated Number of Images | 75,000 | 5,000 |
| Estimated Number of Papers | 4,167 | 277 |
| Average Cost per Paper (USD) | $0.024 | $0.361 |

Note: On average, a single paper, inclusive of Supplementary Information and the main text, contains 18 pages. This was deduced from a sample of 347 papers in this study, which collectively accounted for 6,240 pages. For the Web Interface, a single account priced at $20 per month is capable of processing around 500 images/pages daily. Utilizing five such accounts escalates the monthly image processing capacity to a total of 75,000 images. For the API, each request costs approximately 2000 tokens, equating to $0.02 per image/page, with an average processing time of 5.2 seconds per request.

**References**

(1) Zheng, Z.; Nguyen, H. L.; Hanikel, N.; Li, K. K.-Y.; Zhou, Z.; Ma, T.; Yaghi, O. M., High-yield, green and scalable methods for producing MOF-303 for water harvesting from desert air. *Nat. Protoc.* **2023,** *18*, 136–156.

(2) Wang, B.; Lv, X.-L.; Feng, D.; Xie, L.-H.; Zhang, J.; Li, M.; Xie, Y.; Li, J.-R.; Zhou, H.-C., Highly stable Zr (IV)-based metal–organic frameworks for the detection and removal of antibiotics and organic explosives in water. *J. Am. Chem. Soc.* **2016,** *138* (19), 6204-6216.

(3) Zheng, Z.; Hanikel, N.; Lyu, H.; Yaghi, O. M., Broadly Tunable Atmospheric Water Harvesting in Multivariate Metal–Organic Frameworks. *J. Am. Chem. Soc.* **2022,** *144* (49), 22669-22675.

(4) Wang, X.; Han, X.; Zhang, J.; Wu, X.; Liu, Y.; Cui, Y., Homochiral 2D porous covalent organic frameworks for heterogeneous asymmetric catalysis. *J. Am. Chem. Soc.* **2016,** *138* (38), 12332-12335.

(5) Liang, C.-C.; Shi, Z.-L.; He, C.-T.; Tan, J.; Zhou, H.-D.; Zhou, H.-L.; Lee, Y.; Zhang, Y.-B., Engineering of pore geometry for ultrahigh capacity methane storage in mesoporous metal–organic frameworks. *J. Am. Chem. Soc.* **2017,** *139* (38), 13300-13303.

(6) Zhang, Y.-B.; Furukawa, H.; Ko, N.; Nie, W.; Park, H. J.; Okajima, S.; Cordova, K. E.; Deng, H.; Kim, J.; Yaghi, O. M., Introduction of functionality, selection of topology, and enhancement of gas adsorption in multivariate metal–organic framework-177. *J. Am. Chem. Soc.* **2015,** *137* (7), 2641-2650.

(7) Nguyen, H. L.; Gropp, C.; Hanikel, N.; Möckel, A.; Lund, A.; Yaghi, O. M., Hydrazine-hydrazide-linked covalent organic frameworks for water harvesting. *ACS Cent. Sci.* **2022,** *8* (7), 926-932.

(8) Kim, M.; Cahill, J. F.; Fei, H.; Prather, K. A.; Cohen, S. M., Postsynthetic ligand and cation exchange in robust metal–organic frameworks. *J. Am. Chem. Soc.* **2012,** *134* (43), 18082-18088.

(9) Zheng, Z.; Zhang, O.; Borgs, C.; Chayes, J. T.; Yaghi, O. M., ChatGPT Chemistry Assistant for Text Mining and Prediction of MOF Synthesis. *J. Am. Chem. Soc.* **2023,** *145* (32), 18048–18062.